\documentclass[10pt,twocolumn,letterpaper]{article}

\usepackage[pagenumbers]{cvpr} 

\usepackage{graphicx}
\usepackage{amsmath}
\usepackage{amssymb}
\usepackage{multirow}
\usepackage{booktabs}
\usepackage{tabularx}
\usepackage[dvipsnames,table,xcdraw]{xcolor}
\usepackage{overpic}
\usepackage{colortbl}
\usepackage{makecell}
\usepackage{contour}

\definecolor{cvprblue}{rgb}{0.21,0.49,0.74}
\usepackage[pagebackref,breaklinks,colorlinks,linkcolor=BrickRed,citecolor=cvprblue]{hyperref}

\usepackage{enumitem}
\usepackage{pifont}

\newcommand{\figref}[1]{Fig.~\ref{#1}}
\newcommand{\tabref}[1]{Tab.~\ref{#1}}
\newcommand{\eqnref}[1]{Eqn.~(\ref{#1})}

\newcommand{\secref}[1]{Sec.~\ref{#1}}
\newcommand{\myPara}[1]{\vspace{6pt}\noindent\textbf{#1}}

\newcommand{\jpt}[1]{\textcolor{black}{#1}}
\newcommand{\yyq}[1]{\textcolor{black}{#1}}

\newcommand{\myParaP}[1]{\vspace{.05in}\noindent\textbf{#1}}

\newcommand{\tablestyle}[2]{\setlength{\tabcolsep}{#1}\renewcommand{\arraystretch}{#2}\centering\footnotesize}
\newcommand{\bd}[1]{\textbf{#1}}
\newlength\savewidth

\newcolumntype{x}[1]{>{\centering\arraybackslash}p{#1pt}}

\def\paperID{1592} 
\def\confName{CVPR}
\def\confYear{2024}

\graphicspath{{./figs/}, {./figs/vis/}}

\begin{document}

\title{Multi-Task Dense Prediction via Mixture of Low-Rank Experts}

\author{
  Yuqi Yang$^{1,2}$\thanks{The first two authors contributed equally to this paper. Work was done when Yuqi Yang was an intern at vivo. } ~~ Peng-Tao Jiang$^{{2*}}$ ~~ Qibin Hou$^{1,3}$\thanks{Qibin Hou is the corresponding author. 
  }  ~~ Hao Zhang$^2$ ~~ Jinwei Chen$^2$ ~~ Bo Li$^2$ \\
$^1$VCIP, CS, Nankai University \quad $^2$vivo Mobile Communication Co., Ltd  \quad $^3$NKIARI, Shenzhen Futian\\
{\tt\small $\tt yangyq2000@mail.nankai.edu.cn$,  
$ \tt pt.jiang@vivo.com$,
$ \tt andrewhoux@gmail.com$ }}

\maketitle

\begin{abstract}

Previous multi-task dense prediction methods based on the Mixture of Experts (MoE) have received great performance but they neglect the importance of explicitly modeling the 
global relations among all tasks.
In this paper, we present a novel decoder-focused method 
for multi-task dense prediction, called Mixture-of-Low-Rank-Experts (MLoRE).
To model the global task relationships, MLoRE adds a generic 
convolution path to the original MoE structure, where each task 
feature can go through this path for explicit parameter sharing.
Furthermore, to control the parameters and computational cost brought by the increase in the number of experts,  we take inspiration from LoRA and propose to leverage the low-rank format of a vanilla convolution in the expert network.
Since the low-rank experts have fewer parameters and can be dynamically parameterized into the generic convolution, the parameters and computational cost do not change much with the increase of experts.
Benefiting from this design, we increase the number of experts and 
its reception field to enlarge the representation capacity, 
facilitating multiple dense tasks learning in a unified network.
Extensive experiments on the PASCAL-Context and NYUD-v2 benchmarks show
that our MLoRE achieves superior performance compared to previous 
state-of-the-art methods on all metrics.
Our code is available at \url{https://github.com/YuqiYang213/MLoRE}.
\end{abstract}

\section{Introduction} \label{sec:intro}
Computer vision tasks, such as semantic segmentation \cite{long2015fully,chen2017deeplab,lin2017refinenet,yu2018bisenet} 
and depth estimation \cite{ranftl2021vision,bhat2021adabins}, 
have been significantly facilitated by deep learning techniques.
Each vision task has its own elaborated deep models that usually 
follow a similar pipeline, \ie, feature extraction and prediction.
Besides, some tasks also share relations.
These facts motivate researchers to study Multi-Task Learning (MTL) 
that is able to unify different task models into a single one.
The significant advantage of multi-task learning lies in that it can 
improve training and inference efficiency while keeping commensurate 
performance to each task model.
Due to such an advantage, MTL models have been applied to several 
directions, including autonomous driving \cite{ishihara2021multi,yu2020bdd100k,li2018rethinking} 
and scene understanding \cite{xu2018pad,ye2022inverted}, \etc.

\begin{figure}[t]
    \centering
    \setlength\tabcolsep{1pt}
    \setlength{\abovecaptionskip}{4pt}
    \begin{overpic}[width=0.48\textwidth]{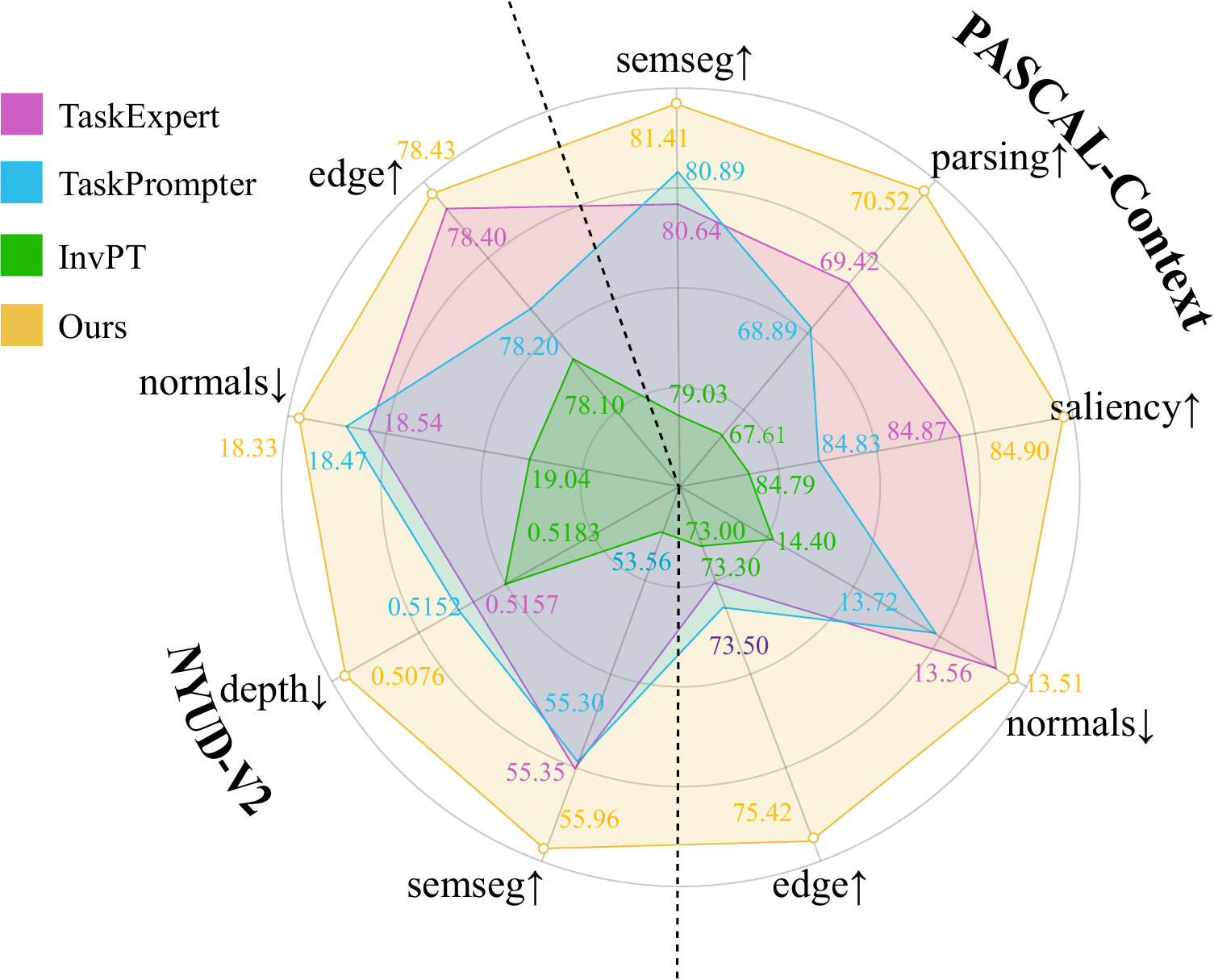}
    \end{overpic}
    \caption{Performance comparison with state-of-the-art methods. 
    Our MLoRE based on the proposed mixture of low-rank experts achieves superior performance on all tasks. $\uparrow$ denotes higher is better.
    $\downarrow$ denotes lower is better.
    }\label{fig:teaser}
\end{figure}

In this paper, we focus on multi-task learning for dense scene understanding, 
in which each task predicts pixel-wise results.
A research line of early works~\cite{misra2016cross,lu2017fully,gao2019nddr,xu2018pad,vandenhende2020mti,bruggemann2021exploring,ye2022inverted} focuses on designing delicate network architectures, including 
encoder-focused and decoder-focused methods.
The encoder-focused methods~\cite{misra2016cross,lu2017fully,gao2019nddr} design hand-crafted modules to share across task-specific encoders to construct task-generic features, while the decoder-focused methods~\cite{xu2018pad,vandenhende2020mti,bruggemann2021exploring,ye2022inverted} propose tailored decoders to learn discriminative task-specific representations and build cross-task relationships.

Unlike the above methods focusing on designing static network architectures, 
some methods \cite{chen2023mod, fan2022m3vit, chen2023adamv} introduce 
the mixture-of-experts (MoE) technique that sheds light on learning dynamic and automatic manner of task specialization and cooperations~\cite{chen2023mod}.
They utilize MoE to design the encoder blocks and dynamically select network paths for different tasks and different samples.
However, compared to these encoder-focused methods, the utilization of MoE in the decoder is less studied, where only Ye \etal \cite{ye2023taskexpert} first 
applied MoE to the decoder recently. 
It decodes task-specific features 
by dynamically assembling task-generic features from different experts and achieves superior performance than previous decoder-focused methods.
This motivates us to delve deep into the MoE-based MTL decoder.

Benefiting from the dynamic routing process, these MoE-based methods 
can significantly improve the diversity in both parameters and features, 
resulting in more discriminative task-specific features. 
However, such a paradigm still has some limitations.
First, although MoE-based methods can build connections in a subset of tasks by sharing the same expert in the dynamic routing process, the chances of expert sharing among all the tasks are very low, potentially 
hindering the router from building global relationships among all tasks.
Nevertheless, the global relationship modeling among all tasks has been 
proven useful for dense MTL in~\cite{ye2022inverted,ye2022taskprompter}. 
Based on this fact, we argue it is essential to explicitly model 
global relationship among all tasks in MoE.
Furthermore, the capacity of the task-generic feature spaces is highly 
related to the number of experts.
As empirically demonstrated in~\cite{fedus2022switch,clark2022unified}, 
increasing the number of experts 
has the potential to facilitate multi-task learning in a unified network.
However, for existing dense multi-task learning, adding more expert networks 
will introduce more parameters and computational costs, which is a 
heavy burden for the whole model.

\begin{table}[t]
    \centering
    \small
    \renewcommand{\arraystretch}{1.05}
    \caption{Parameters and FLOPs of different settings in the standard MoE and 
    the proposed MLoRE.  The left setting presents the number of experts 
    and the kernel size of the convolutions in the expert network.
    }\label{tab:moe_param_comp}
    \vspace{-7pt}
    \setlength{\tabcolsep}{2.0mm}{
    \begin{tabular}{ccccc} \toprule 
        \multirow{2}*{\bd{Settings}}  & \multicolumn{2}{c}{\bd{Params (M)}}  & \multicolumn{2}{c}{\bd{FLOPs (G)}}   \\ \cmidrule(lr){2-3}\cmidrule(lr){4-5}
                   & MoE   &  MLoRE   & MoE   &  MLoRE  \\ \toprule 
        5 experts,  [1$\times$1, 1$\times$1] & 3.1  & 1.2  & 3.00  & 1.49\\
        10 experts, [1$\times$1, 1$\times$1] & 4.7  & 1.6  & 4.49  & 1.58\\
        15 experts, [1$\times$1, 1$\times$1] & 6.3  & 1.9  & 5.99  & 1.66 \\ \hline
        5 experts,  [3$\times$3, 1$\times$1] & 14.9 & 3.4 & 14.24 & 7.12\\
        10 experts, [3$\times$3, 1$\times$1] & 22.4 & 4.7 & 21.37 & 7.21\\
        15 experts, [3$\times$3, 1$\times$1] & 29.9 & 6.0 & 28.49 & 7.29\\
        \bottomrule
        \end{tabular} }
\end{table}

To address the above issues, we propose a novel decoder-focused approach, 
which we call it Mixture of Low-Rank Experts (MLoRE).
The core idea of the MLoRE framework is to explicitly model 
global relationships among all tasks in the MoE and free the MoE from heavily 
dense computation burdens when increasing the number of experts 
to enlarge the task-generic feature spaces and context.
To address the first issue, MLoRE builds upon the basic MoE structure 
and introduces a task-sharing generic convolution path 
that parallels with the MoE.
Specifically, the backbone features are first projected to 
different task features and then all of them are fed 
into the generic convolution path and the original MoE's 
expert networks.
By simply sharing the same generic path among all tasks, 
different tasks can be globally correlated with each other.
Furthermore, to enhance the discrimination of task-specific 
features, we exclude some experts from the dynamic routing process 
and enable them to serve specific tasks.

To increase the number of experts while not bringing too many 
parameters and FLOPs, we take inspiration from LoRA that the basic models 
adapted to different tasks only needs low-rank weight updates.
To be specific, we transform the expert networks of the MoE to 
the different low-rank formats of a vanilla convolution, 
which saves more than 60\% of the parameters compared 
to the standard MoE module, as shown in \tabref{tab:moe_param_comp}.
In addition, to control the computational cost brought by the increasing 
number of expert networks, we do not use \yyq{any non-linear} activation functions 
in all the expert networks and the generic convolution path, 
which enables re-parameterization during inference. 
Through re-parameterization, the knowledge of experts can be injected 
into the generic convolution path, reducing the computational cost 
for dense tasks.
\yyq{To our knowledge, we are the first to use linear experts in MoE for multi-task dense prediction.}
To verify the effectiveness of \yyq{our} method, we conduct 
comprehensive experiments on the PASCAL-Context and NYUD-v2 datasets.
Our method achieves new performance records on 
all tasks as shown in~\figref{fig:teaser}.

In summary, our contributions are three-fold:
\begin{itemize}
    \item We analyze the issues of MoE when applied in multi-task learning and propose a novel decoder-focused framework, MLoRE, which can explicitly model global relationships among all tasks and enlarge the capacity of feature representations without increasing the model size too much.

    \item We introduce a simple task-sharing generic path to the MoE structure and propose linear and low-rank expert networks inspired by LoRA. The generic convolution path and low-rank expert paths can be linearly combined, enabling re-parameterization at inference.
    
    \item Experiments on PASCAL-Context and NYUDv2 show 
    that the proposed method clearly outperforms previous state-of-the-art MTL 
    methods on all tasks.
\end{itemize}

\section{Related Work} \label{sec:related}
\subsection{Dense Multi-Task Learning}
In computer vision, multi-task learning (MTL) for dense prediction tasks 
has been widely studied.
The previous methods can be divided into two categories, including 
optimization-based methods and architecture-based methods~\cite{vandenhende2021multi}.
Optimization methods~\cite{kendall2018multi, chen2018gradnorm, guo2018dynamic,zhao2018modulation, chen2020just} facilitate MTL by utilizing 
different strategies to balance the influence of each task. 
Architecture-based methods aim to design a unified deep 
network for \yyq{MTL}.
They can be further classified into two categories, 
encoder-focused methods, and decoder-focused methods.
\emph{$($i$)$}  Encoder-focused methods 
\cite{ruder2019latent,lu2017fully,bruggemann2020automated,guo2020learning} 
design multi-task backbones to extract features adapted to different tasks.
Typical methods include cross-stitch networks \cite{misra2016cross}, 
neural discriminative dimensionality reduction networks \cite{gao2019nddr}, 
multi-task attention networks \cite{liu2019end}, branched networks \cite{vandenhende2019branched}, 
and mixture-of-expert networks \cite{fan2022m3vit,chen2023mod}.
\emph{$($ii$)$} Decoder-focused methods \cite{xu2018pad,vandenhende2020mti,bruggemann2021exploring,ye2022inverted,ye2022taskprompter,ye2023taskexpert,zhang2019pattern,zhang2018joint,zhou2020pattern} 
share the same backbone and design delicate heads 
to extract task-specific features for each task and 
cross-task relationships.
The advantage of decoder-focused methods lies in that 
they can benefit from powerful off-the-shelf vision backbones, 
such as DINOv2 \cite{oquab2023dinov2}.
Our method also falls into the decoder-focused category and 
studies how to produce task-specific features with the MoE technique.

\subsection{Mixture-of-Experts}

Mixture-of-Experts (MoE)~\cite{jacobs1993learning,jacobs1991adaptive} 
learns multiple expert networks and a router network that controls 
the probability of each expert contributing to the final output.
This technique is also used in multi-task learning, 
which can better adapt to the data diversity.
Different expert networks learn different discriminative features.
The router network learns hard/soft task-specific coefficients 
to dynamically assemble discriminative features for each task.
Prior MoE-based multi-task methods \cite{fan2022m3vit,chen2023mod, chen2023adamv} are mainly encoder-focused methods. 
They introduce MoE into the backbone blocks to sparsely activate 
different paths for different tasks in the inference stage.
Recently, Ye \etal \cite{ye2023taskexpert} first introduced the 
MoE technique to the decoder. 
They utilize the spatial context-aware gates to combine each pixel 
of the features from different expert networks.
The above methods decompose the backbone feature into multiple generic feature spaces 
and assemble discriminative task-specific features from them.
Unlike the above MoE-based MTL method, our method first builds global relationships 
among all tasks explicitly in the MoE structure rather than leaving 
this work implicitly done by the task-specific routers.
Moreover, our proposed low-rank experts give MoE better efficiency compared to the naive MoE and this gap gradually gets larger as the number of experts increases.

\subsection{Low-Rank Structure}
The low-rank structure is often used in deep learning for its efficiency~\cite{udell2016generalized, idelbayev2020low, yang2016trace, su2015multi}.
Recently, many methods in parameter-efficient adaptation \cite{hu2021lora, liu2022polyhistor, jie2023fact, sung2022vl}, such as LoRA \cite{hu2021lora}, utilize the low-rank structure and have shown impressive results.
LoRA takes inspiration from Aghajanyan \etal \cite{aghajanyan2020intrinsic} that the difference 
in weights between the pre-trained model and the adapted model resides on low intrinsic rank.
It learns an extra low-rank matrix instead of tuning the whole layer for adaptation.
More related to our work, earlier MTL methods~\cite{yang2016trace, su2015multi} utilize 
low-rank structure to model task-generic features and generate task-specific features 
by linear combination.
Unlike them, our method utilizes the low-rank structure to control the computation budget 
when increasing the number of experts in MoE.

\begin{figure*}[t]
    \centering
    \setlength\tabcolsep{1pt}
    \begin{overpic}[width=\textwidth]{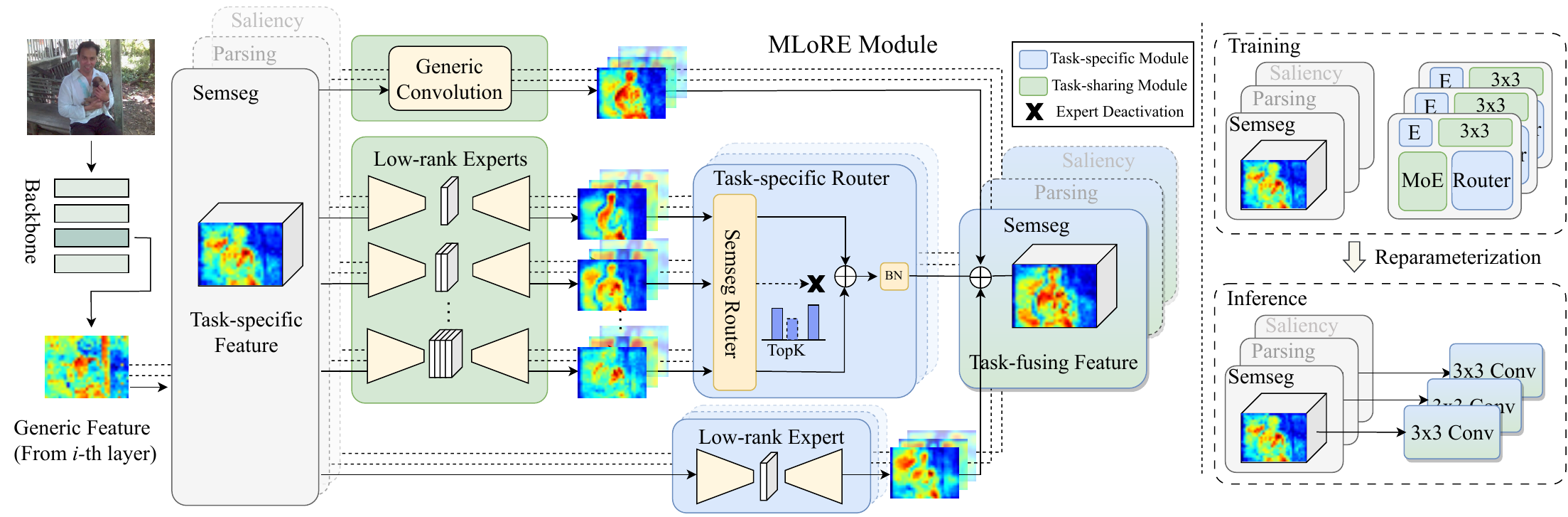}
    \end{overpic}
    \caption{Overall framework of the proposed method. 
    The MLoRE modules are equipped at different layers, where the backbone features 
    from different layers are fed into the MLoRE modules, respectively.
    At each selected layer, the backbone feature is first projected 
    to different task features and then sent to the task-sharing convolution, 
    task-sharing low-rank expert networks followed by the task-specific router network 
    and task-specific low-rank expert networks.
    The outputs of these branches are accumulated to generate task-specific features.
    At each selected layer, we stack two MLoRE modules.
    }\label{fig:pipeline}
\end{figure*} 

\section{Method}

\subsection{Overall Framework}
The overall framework follows the multi-scale architecture 
as in previous works \cite{ye2023taskexpert,ye2022taskprompter}.
We utilize an off-the-shelf vision transformer (ViT) as the encoder 
and collect multi-scale features from different layers.
Formally, given an input image $\mathbf{I}$ and a vision transformer $\mathcal{F}$, 
we can attain multi-scale feature sets 	
$\left \{ \mathbf{X}^\emph{l} = \mathcal{F}^\emph{l}(\mathbf{I})\right \}$ 
from different layers, 
where $\mathbf{X}^l \in \mathbb{R}^{L \times C}$. 
Here, $l$, $L = H\times W$, and $C$ denote the layer index, the number of patches, 
and the feature dimension, respectively.
$\mathcal{F}^\emph{l}(\mathbf{I})$ is the output features of the $l$-th transformer layer.
The multi-scale features are fed into the decoder which includes 
two stacked MLoRE modules at each scale.
For each task, the output features of MLoRE from different scales 
are concatenated together to generate the final task features 
for dense predictions.

\subsection{Preliminaries: Mixture of Experts}

Before describing the details of the proposed Mixture-of-Low-Rank-Experts (MLoRE) module, we first introduce the basic form of MoE $f_{moe}(\cdot)$.
Formally, suppose MoE contains $N$ experts and $T$ router networks, 
denoted as $\mathbb{E} = \left\{ E_1, E_2, ..., E_N \right \}$ and 
$\mathbb{G} = \left\{ G_1, G_2, ..., G_T \right \}$, respectively. 
$N$ and $T$ are the number of experts and the number of tasks, respectively.
The backbone feature $\mathbf{X}^\emph{l}$ from the $l$-th layer is fed 
into the networks of $N$ experts and $T$ router networks, respectively.
For convenience, the superscript $l$ is omitted in the following.
For the $n$-th expert, the  discriminative output feature is generated by 
$\mathbf{X}_n = E_n(\mathbf{X})$.
In the meanwhile, MoE learns gates from the task-specific router networks 
for different tasks. 
For task $t$, the gate value for each expert generated by the router network can be represented as $\mathbf{g}^t = G_t(\mathbf{X})$, 
where $\mathbf{g}^t \in \mathbb{R}^{N}$.
Finally, MoE generates task-specific feature $\mathbf{S}^t$ for task $t$ 
by utilizing gates to combine the expert features, which can be formulated as
\begin{equation}
    \mathbf{S}^t = f_{moe}(\mathbf{X}) = \sum_{n=1}^N \mathbf{g}^t_n \mathbf{X}_n .
 \end{equation}
For M$^3$ViT \cite{fan2022m3vit} and Mod-Squad \cite{chen2023mod}, 
they deactivate some experts according to \yyq{the corresponding} gate values for one inference 
and select the top-$k$ experts. 
The task-specific features are used to make predictions for each task.

The advantages of MoE lie in that it enables the dynamic encoding of features 
for each sample and each task and increases feature encoding diversity 
via multiple experts.
However, when applying the MoE technique to build an MTL decoder, 
we find that it struggles with building global task relationships.
Moreover, when increasing the number of experts to enlarge the 
capacity of feature representations and the context of the expert network, 
the parameters and computational cost increase accordingly.
To address the above issues, we propose the Mixture-of-Low-Rank-Expert 
(MLoRE) module.

\subsection{Mixture of Low-Rank Experts}
The overall pipeline of the proposed Mixture-of-Low-Rank-Expert (MLoRE) module is illustrated in \figref{fig:pipeline}.
Considering building cross-task relations across the task-specific features,
we first project the backbone features into task features 
using several lightweight convolutional layers.
Then, the feature of each task is delivered to the task-sharing 
generic path and multiple task-sharing expert networks with 
different low-rank convolutions.
The low-rank experts are selected according to the predictions 
of the task-specific router network for each task.
Furthermore, except for building task-specific features based on task-specific routers, 
we introduce additional task-specific low-rank expert networks to assist in 
building more discriminative task-specific features.
The features from the task-sharing generic path, task-sharing 
low-rank expert networks selected by task-specific routers and the task-specific 
expert networks are summed up to generate discriminative task-specific features.
We introduce linearity into the MLoRE module and do not utilize any activation 
function in all paths, enabling re-parameterization to reduce computational cost.

To be specific, the backbone feature $\mathbf{X}$ at the $l$-th layer 
is first projected to multiple task splits corresponding to each task using $1\times1$ convolutions, which can be denoted as
$\left\{ \mathbf{X}^t = f_{t, 1\times 1}(\mathbf{X}), t \in [1, \cdots, T] \right\}$, 
where $\mathbf{X}^t \in \mathbb{R}^{C\times H \times W}$.
Then, the task feature is sent to three paths, \ie, task-sharing generic 
path $f_{g}(\cdot)$, task-sharing low-rank expert path $f_{lre}(\cdot)$ 
with task-specific router network $f^t_{sr}(\cdot)$, and task-specific low-rank 
expert path $f_{se}(\cdot)$.
The output task-specific feature $\mathbf{S}^t$ is 
obtained by
\begin{equation} \label{eqn:total}
    \mathbf{S}^t = f_g(\mathbf{X}^t) + f^t_{sr} \left(f_{lre}(\mathbf{X}^t) \right) + f^t_{se}(\mathbf{X}^t).
\end{equation}
In the following, we introduce the network details of all these paths 
in the MLoRE module.

\myParaP{Task-sharing generic path}
contains a simple 3$\times$3 convolutional layer 
with a weight matrix $\mathbf{W}_g \in \mathbb{R}^{3\times3\times C \times C}$ 
and a bias matrix $\mathbf{b}_g \in \mathbb{R}^{C}$.
As all task features will go through this generic convolution, 
it will be optimized by the gradients of different tasks simultaneously, 
which can help extract common features among all tasks.
During the training process, we stop the gradients of this path for further back-propagation.
\yyq{The gradient is back-propagated through the other two paths.}
We found such a simple operation can better ease the optimization process and well solve the gradient conflicts.
Experimental results in \secref{sec:ablation} show that the simple 
task-sharing generic path can bring performance improvements on all tasks, demonstrating the effectiveness of the idea of explicitly building the relationship across all tasks from a global perspective.

\myParaP{Task-sharing low-rank expert path.} 
We take inspiration from LoRA \cite{hu2021lora} 
and adopt low-rank convolution which is a low-rank format 
of a vanilla convolution.
Each task-sharing expert network shares a similar structure consisting of a $3\times3$ convolution and a $1\times1$ convolution.
The weights and bias of all task-sharing expert networks can be formulated as 
\yyq{
$\left\{ \mbox{\bd{W}}^n_{lreb}, \mbox{\bd{b}}^n_{lreb}, \mbox{\bd{W}}^n_{lrea}, 
\mbox{\bd{b}}^n_{lrea} | n \in [1,..., N] \right\}$, 
where $\mbox{\bd{W}}^n_{lreb}\in \mathbb{R}^{3\times3\times C \times r_n},  
\mbox{\bd{b}}^n_{lreb}\in \mathbb{R}^{r_n},
\mbox{\bd{W}}^n_{lrea} \in \mathbb{R}^{1\times1\times r_n \times C}$}
, and 
\yyq{$\mbox{\bd{b}}^n_{lrea}\in \mathbb{R}^{C}$ ($r_n\ll C$).}
$r_n$ denotes the rank for the $n$-th expert network. 
In our method, $r_n$ for different expert networks is different, aiming to improve the diversity of parameters and features.
\jpt{For each task, the task-specific router network $f^t_{sr}(\cdot)$ learns gate values for these experts and activate 
the top-$k$ experts according to the gate values.}
The output features of all the activated experts are 
summed up and then sent to a BatchNorm layer to generate 
task-specific features. 
The BatchNorm layer contains four parameters, including the accumulated 
channel-wise mean $\mathbf{\mu} \in \mathbb{R}^C$, the accumulated 
channel-wise standard deviation $\mathbf{\sigma} \in \mathbb{R}^C$, 
the scaling factor $\gamma \in \mathbb{R}^C$ and the $\beta \in \mathbb{R}^C$, 
respectively.

\myParaP{Task-specific low-rank expert path} includes $T$ specific 
expert networks and each processes one task feature.
For each specific expert network, we utilize a similar structure 
with task-sharing expert path, which contains a $3\times3$ convolution 
with a weight matrix $\mathbf{W}^t_{seb} \in \mathbb{R}^{3\times3\times C \times R}$ 
and a bias matrix $\mathbf{b}^t_{seb} \in \mathbb{R}^{R}$, followed by a $1\times1$ convolution with a weight matrix 
$\mathbf{W}^t_{sea} \in \mathbb{R}^{1\times1\times R \times C}$ and 
a bias matrix $\mathbf{b}^t_{sea} \in \mathbb{R}^{C}$. 
$R$ denotes the rank number ($R \ll C$).
The task-specific expert path can enhance the \yyq{distinctiveness}
of the task-specific features, which will be verified in the experiments.

\myParaP{Router network.} 
As shown in \figref{fig:pipeline}, to generate task-specific features 
from the task-sharing low-rank expert path, 
we learn task-specific router networks to generate gate values 
for each expert 
\yyq{and utilize them as the weight of the linear combination of feature output from different experts.}
The router network for each task is usually the simple linear layers 
followed by an average pooling layer and a prediction layer.
Specifically, the router network \yyq{$f^t_{sr}(\cdot)$} for the $t$-th task 
is designed as follows.
\yyq{Our router network takes the task-specific features $\mathbf{X}^t \in \mathbb{R}^{C\times H \times W}$ as input and }
feed them into two consecutive 1$\times$1 convolutions, mapping the channel dimension from $C$ to $C/4$, followed by a global pooling layer.
The output is a global feature vector $\mathbf{X}_f \in \mathbb{R}^{\frac{C}{4}}$.

In addition, inspired by previous works \cite{hou2021coordinate,hou2020strip} 
showing that positional information is also important for modeling long-range spatial context, we introduce another parallel position-aware branch.
Similarly, it consists of two linear layers.
The first linear layer shrinks the feature along the spatial dimension, mapping the shape from $\mathbb{R}^{C \times HW}$ to $\mathbb{R}^{C \times 1}$, which will then be transformed to $\mathbb{R}^{\frac{C}{4}}$ via the second linear layer.
The output feature vectors of these two branches are concatenated 
along the final dimension and then sent to the final prediction layer 
followed by a Softmax function to produce the gate values \yyq{$g_t$} for each expert.

\myParaP{Re-parameterization during inference.} 
We introduce linearity into the MLoRE module by removing 
all activation functions, enabling the parameters of all paths to 
reparameterize to a simple 3$\times$3 convolution for each task 
at inference. 
We first parameterize the task-sharing low-rank expert path and 
then parameterize the parameters of all paths.
According to \cite{ding2021diverse}, 
the weight and bias matrices in the task-sharing expert path can be combined and formulated as :
\begin{align}
 \mathbf{W}^t_{lre} &= \mathfrak{B} (\frac{\gamma}{\sigma}) \sum_{k \in \mathbb{K}_t} \mathbf{g}^t_k \mathbf{W}^k_{lreb} \mathbf{W}^k_{lrea}, \\
 \mathbf{b}^t_{lre} &= \frac{\gamma}{\sigma} ( \sum_{k \in \mathbb{K}_t} \mathbf{g}^t_k (\mathbf{b}^k_{lreb} \mathbf{W}^k_{lrea}  + \mathbf{b}^k_{lrea}) - \mu ) + \beta,
\end{align}
where $\mathfrak{B}$ denotes the broadcast operation and $\mathbb{K}_t$ denotes 
the index set of the activated experts selected by the router network for task $t$.
\jpt{$\mathbf{g}^t_k$ is the $k$-th gate value predicted by the router.}
At the inference time, the weight matrix and bias matrix of 
all these three paths can be re-parameterized as 
\begin{align}
\mathbf{W}^t_r &= \mathbf{W}_g + \mathbf{W}^t_{sr} + \mathbf{W}^t_{seb} \mathbf{W}^t_{sea}, \\
\mathbf{b}^t_r & = \mathbf{b}_g + \mathbf{b}^t_{sr} + \mathbf{b}^t_{seb}\mathbf{W}^t_{sea} + \mathbf{b}^t_{sea}.
\end{align}
$\mathbf{W}^t_r$ and $\mathbf{b}^t_r$ are the weight and bias 
of the re-parameterized convolution.
Thus, \eqnref{eqn:total} can be reformulated as 
\begin{equation}
    \mathbf{S}^t = \mathbf{X}^t \circledast \mathbf{W}^t_r + \mathfrak{B}(\mathbf{b}^t_r),
\end{equation} 
where $\circledast$ denotes the convolution operation and 
$\mathbf{b}^t_r$ is with the same shape as $\mathbf{X}^t$ via broadcast.

\section{Experiments}

\subsection{Experimental Settings}
\myParaP{Datasets.}
To demonstrate the effectiveness of our method, we evaluate the performance of our method 
on two popular multi-task datasets, including PASCAL-Context \cite{chen2014detect} 
and NYUD-v2 \cite{silberman2012indoor}. 
PASCAL-Context \cite{chen2014detect} contains high-quality annotations 
of several tasks, including semantic segmentation, human parsing, 
saliency detection, surface normals, and object boundary detection.
There are 4,998 training images and 5,105 test images in this dataset.
NYUDv2 \cite{silberman2012indoor} also provides high-quality multi-task annotations, 
including semantic segmentation, monocular depth estimation, surface normals, 
and object boundary detection.
This dataset contains 795 training images and 654 test images.

\myParaP{Evaluation metrics.}
\yyq{We introduce the evaluation metrics 
for the tasks mentioned above.}
Following previous multi-task works \cite{ye2022inverted,ye2022taskprompter}, 
the mean intersection-over-union (mIoU) is used to evaluate 
semantic segmentation and human parsing.
The root mean square error (RMSE) is used to evaluate 
the accuracy of monocular depth estimation.
Saliency detection \yyq{uses} the maximum F-measure (maxF).
Surface normal and object boundary detection adopt 
the mean error (mErr) and the optimal-dataset-scale F-measure (odsF) 
as the evaluation metrics, respectively.
In total, we evaluate the MTL gain $\Delta_m$ 
across all tasks, following \cite{maninis2019attentive}.

\begin{table}[t]
    \small
    \centering
    \caption{Ablation study on different components in MLoRE on the PASCAL-Context dataset. 
    Each row adds an extra setting to the above row. 
    MoE: the standard mixture-of-experts structure;
    LoRE: task-sharing low-rank expert path;
    GC: task-sharing generic convolution path; 
    SPE: task-specific expert path.
    $\uparrow$ denotes higher is better.
    $\downarrow$ denotes lower is better.
    }
    \label{tab:ablation_component}
    \renewcommand{\arraystretch}{1.05}
    \setlength\tabcolsep{0.15mm}
    \vspace{-8pt}
    \begin{tabular}{l|cccccc|cc} \hline 
    \bd{Settings} & \thead{\bd{Semseg} \\ mIoU $\uparrow$ } & \thead{\bd{Parsing} \\ mIoU $\uparrow$ } 
    & \thead{\bd{Sal.} \\ maxF $\uparrow$ } & \thead{\bd{Normal} \\ mErr $\downarrow$ } 
    & \thead{\bd{Bound.} \\ odsF $\uparrow$ } &  \thead{\bd{MTL} \\ $\Delta_m$ $\uparrow$ } 
    & \thead{\bd{FLOPs} \\ (G)}  & \thead{\bd{\#Param} \\ (M)}   \\ 
       \hline 
        Baseline  & 77.38 & 65.15 & 85.08 &13.79  &  69.87  & -3.41   & 391 & 115  \\ 
        $+$ MoE   & 78.56 & 66.78 & 85.18 & 13.57 &  73.91  & -1.20   & 1834& 676  \\ 
        \hline 
        Baseline  \\
        $+$ LoRE  & 78.38 & 66.21 & 85.15 &  13.71 & 73.53  & -1.71   & 568 & 213   \\
        $+$ GC    & 79.25 & 67.43 & 85.20 & 13.70  & 74.38  & -0.88   & 568 & 243   \\
        $+$ SPE   & 79.26 & 67.82 & 85.31 & 13.65 & 74.69   & -0.58   & 568 & 259   \\ \hline 
    \end{tabular} 
\end{table}

\myParaP{Training settings.} 
We utilize ViT-large \cite{dosovitskiy2020image} as the backbone. 
\yyq{The channel number of the decoder is set to 384.}
For the ablation study,  the ViT-base network is set as the backbone. 
Following the previous work \cite{ye2022taskprompter}, the proposed MTL 
network is trained for 40,000 iterations with a batch size of 4 on 
both the two datasets.
The optimizer and the loss functions for different tasks follow 
the previous work \cite{ye2022taskprompter}.

\subsection{Ablation Study} \label{sec:ablation}
In this subsection, we conduct extensive experiments to demonstrate 
the effectiveness of different components and find the best settings 
of different hyper-parameters.
All the ablation experiments are conducted based on the ViT-based backbone 
if not specified otherwise.
The baseline is built upon the ViT-base backbone with 12 layers, where 
the backbone features from the 3-$rd$, 6-$th$, 9-$th$, and 12-$th$ layers 
are utilized as the multi-scale features, each of which is followed 
by a linear layer to project the channel dimension to the output channels 
for each task.

\myParaP{Effectiveness of different components.} 
We first conduct experiments to verify the effectiveness of different 
components of the MLoRE module.
The quantitative results are shown in \tabref{tab:ablation_component}.
We first examine the performance of the baseline with the standard MoE 
and the parameter size and FLOPs of their model.  
The expert networks in the standard MoE (15 experts) are similar to ours, 
each of which consists of a 3$\times$3 convolution and a 1$\times$1 convolution 
with ReLU among them.
When adding the MoE to the baseline, we observe the MTL gain can be improved, 
but the parameters and FLOPs also expand about 5 times and 4 times, 
which is a heavy burden for the whole network. 
When adding the low-rank expert networks (LoRE) to the baseline, 
the performance is also improved, but the parameters and FLOPs are only 
1/3 and 1/3 of the MoE-based model.
When the low-rank property is applied to the expert networks,   
the parameter size is reduced several times.
When further introducing linearity in the expert networks by removing 
all activation functions, the computations are saved by re-parameterizing all experts to a single convolution. 

Furthermore, we also emphasize the importance of explicitly building 
global task correlations in the MoE and introduce the task-sharing generic path 
to achieve this goal.
It can be seen that adding the task-sharing generic path to LoRE 
can further improve the performance and outperform the baseline with MoE 
on most metrics, which \yyq{proves} the effectiveness of modeling the 
global relationships among all tasks.  
In addition, adding a task-specific low-rank expert for each task also 
improves the performance, which demonstrates that the specialized 
expert can enhance the discrimination of task-specific features.
We empirically set the rank of the task-specific low-rank expert
to 64 in this paper.

\begin{figure}[t]
    \centering
    \setlength{\abovecaptionskip}{4pt}
    \setlength\tabcolsep{1pt}
    \begin{overpic}[width=0.48\textwidth]{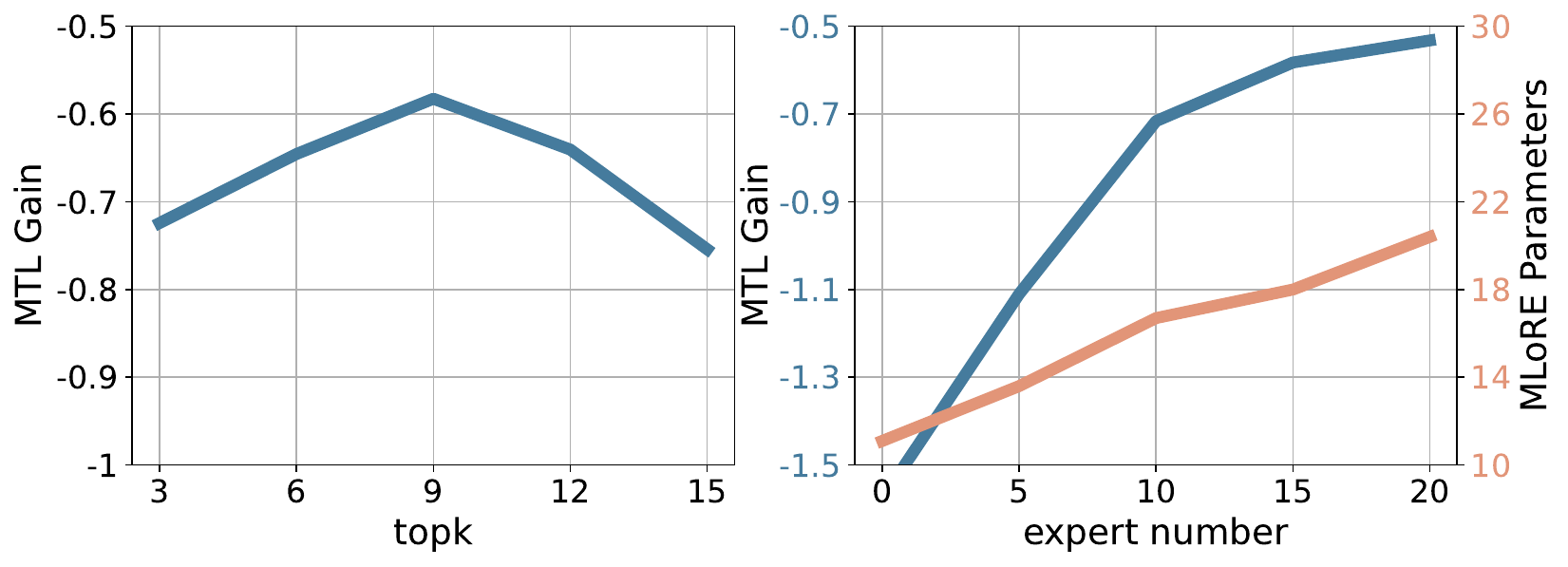}
    \end{overpic}
    \caption{Ablation study on the number of experts $N$ and the number 
    of activated experts $K$. In the right figure, we also present the parameter change 
    of the MLoRE module with the increase in the number of experts.
    }\label{fig:ablation_n_k}
\end{figure} 

\myParaP{Number of task-sharing low-rank experts and top-$k$ selection.}
We ablate the number of low-rank experts in the MLoRE module and 
top-$k$ selection of the experts by the task-specific router networks.
We first fix one and ablate another parameter to study their impact 
on multi-task performance.
As shown in \figref{fig:ablation_n_k}, when increasing the number of experts, 
the MTL gain of the model is significantly improved and allows us to achieve the best performance when the number of experts is 15.
Further increasing the number of experts, we do not observe obvious 
performance gain.
Besides, when fixing the number of experts to 15, we ablate the ratio of 
the activated experts.
We observe that activating 60\% experts for each task is the best choice 
in our experiments.
When selecting all expert networks, the performance decreases by a large margin, 
which reflects the importance of sparsity for feature discrimination.

\begin{figure*} [t]
  \begin{subfigure}{.53\textwidth}
    \centering
    \includegraphics[width=1\linewidth]{ablation_vis_route_te_1_3.png}
\end{subfigure}
  \begin{subfigure}{.45\textwidth}
    \centering
    \includegraphics[width=1\linewidth]{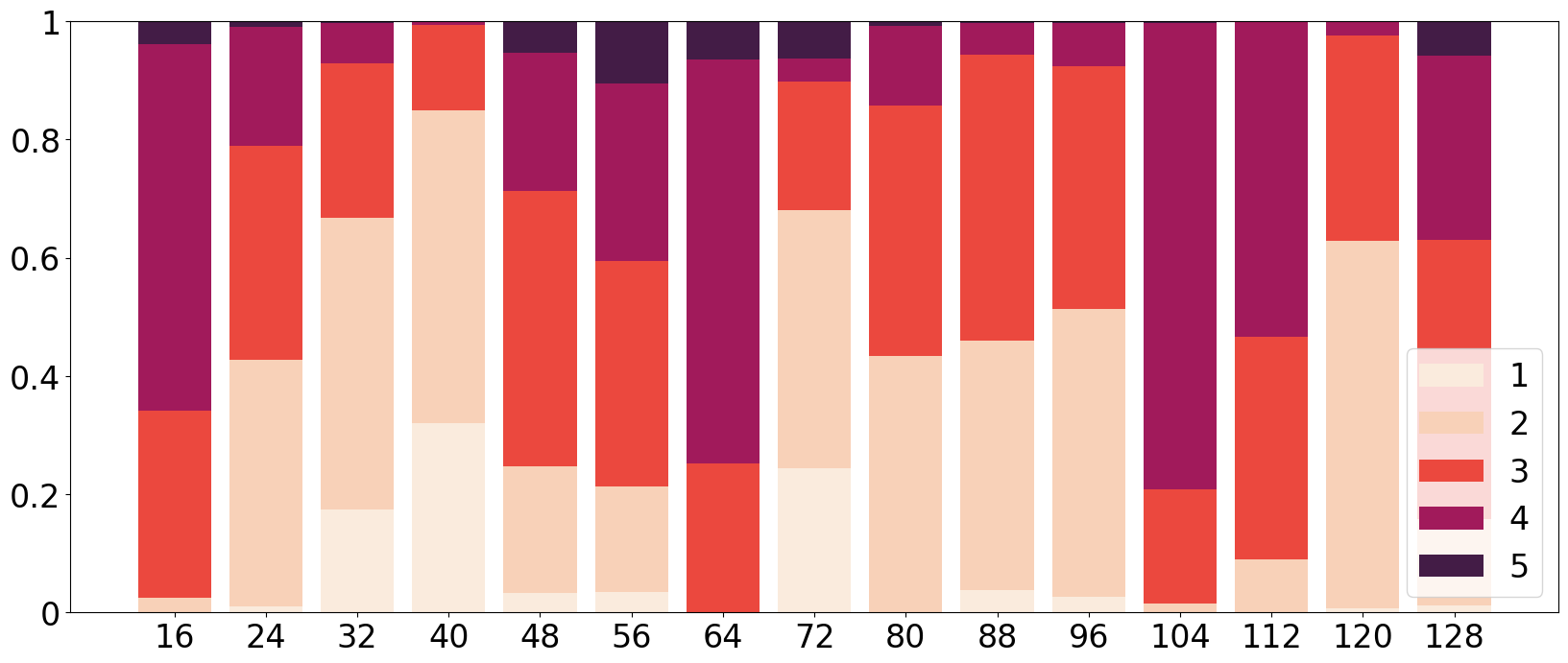}
\end{subfigure}
  \vspace{-5pt}
\caption{
    (a) The relations between tasks and low-rank experts. 
    (b) The ratio of an expert activated by different numbers of tasks in the MLoRE module without the task-sharing generic path.
    We can see that without the task-sharing generic path, there is only a few experts can be activated by all five tasks.
    Horizontal coordinates represent the ranks of different experts.
  }\label{fig:vis_relation}
  \vspace{-10pt}
\end{figure*}

\begin{table}[t]
    \small
    \centering
    \caption{Ablation on the rank setting in the task-sharing low-rank expert path of MLoRE 
    on PASCAL-Context dataset. }
    \label{tab:ablation_rank}
    \renewcommand{\arraystretch}{1.1}
    \setlength\tabcolsep{0.7mm}
    \vspace{-10pt}
    \begin{tabular}{c|ccccc|c} \hline 
    \thead{\bd{Min/Max} \\ \bd{Rank}} & \thead{\bd{Semseg} \\ mIoU $\uparrow$ } & \thead{\bd{Parsing} \\ mIoU $\uparrow$ } & \thead{\bd{Saliency} \\ maxF $\uparrow$ } 
    & \thead{\bd{Normal} \\ mErr $\downarrow$ } & \thead{\bd{Boundary} \\ odsF $\uparrow$ }  & \thead{\bd{MTL} \\ $\Delta_m$ $\uparrow$} \\ 
       \hline %
        16/16   & 78.84 & 68.01 & 85.32 & 13.69 & 74.39 & -0.77 \\ 
        16/128  & 79.26 & 67.82 & 85.30 & 13.65 & 74.69 & -0.58 \\
        128/128 & 78.79 & 66.98 & 85.35 & 13.67 & 74.29 & -1.07 \\
    \hline 
    \end{tabular}  
\end{table}

\myParaP{Rank number setting.}
The expert networks utilize the low-rank format of a 
vanilla 3$\times$3 convolution with 640 output channels. 
The rank $r$ of different expert networks also plays an important role.
We study different settings, including 1) all experts with the same rank number 16, 
2) all experts with the rank number 128, 
3) all experts with the rank numbers from 16 to 128 with an increased step of 8. 
As shown in \tabref{tab:ablation_rank}, it can be seen that 
selecting different rank numbers for expert networks achieves the best MTL gain.
We analyze the expert networks with different rank numbers that can 
bring more feature diversity than the same rank numbers, 
which is more useful for assembling task-specific features, \yyq{and utilize this as our settings.}

\myParaP{The relation visualizations} between tasks and 
the low-rank experts are shown in \figref{fig:vis_relation}(a).
We count the relations from the second MLoRE module at the last stage 
and calculate the activation ratio of 
every expert selected by different tasks on the entire dataset.
It can be seen that experts with different ranks tend to 
learn different subsets of tasks.
Specifically, for this module, experts with lower ranks tend to learn shared knowledge 
for 3-4 related tasks, while experts with higher ranks tend to specialize in 1-2 tasks.
Moreover, we also show the ratio of different experts to be activated 
by different numbers of tasks when the task-sharing generic path is not added 
to the MLoRE module in \figref{fig:vis_relation}(b).
It can be seen that in a fully dynamic manner, all these experts are seldom 
or even never activated by all the tasks in one sample.
This proves the fact that almost no expert can learn global relationships 
across all tasks when utilizing the MoE in the decoder directly.
This phenomenon strongly supports the necessity of the task-sharing generic path 
for explicitly modeling global task relationships.

\begin{table}[t]
    \small
    \centering
    \renewcommand{\arraystretch}{1.1}
    \setlength\tabcolsep{0.5mm}
    \setlength{\abovecaptionskip}{4pt}
    \caption{Ablation on the setting of the router network. 
    \bd{basic}: the basic router network.
    \bd{pos.}: the positional-aware router network.
    \bd{w/o sample-dep}: the input is the learnable parameters.
    }
    \label{tab:ablation_router}
    \begin{tabular}{c|ccccc|c} \hline
    \thead{\bd{Router} \\ \bd{Variant}} & \thead{\bd{Semseg} \\ mIoU $\uparrow$ } & \thead{\bd{Parsing} \\ mIoU $\uparrow$ } & \thead{\bd{Saliency} \\ maxF $\uparrow$ } 
    & \thead{\bd{Normal} \\ mErr $\downarrow$ } & \thead{\bd{Boundary} \\ odsF $\uparrow$ }  & \thead{\bd{MTL} \\ $\Delta_m$ $\uparrow$} \\ 
       \hline 
        basic        & 79.15 & 67.40 & 85.21 & 13.58 & 74.34 & -0.75 \\
        basic$+$pos.& 79.26 & 67.82 & 85.31 & 13.65 & 74.69 & -0.58 \\
        only pos.   & 79.10 & 67.76 & 85.11 & 13.71 & 74.51 & -0.82 \\
        \hline
        basic          & 79.15 & 67.40 & 85.21 & 13.58 & 74.34 & -0.75 \\
        w/o sample-dep & 78.86 & 67.38 & 85.41 & 13.65 & 74.25 & -0.91 \\
        \hline 
    \end{tabular}  
\end{table}

\begin{table*}[t]
    \small
    \centering
    \setlength{\abovecaptionskip}{4pt}
    \renewcommand{\arraystretch}{1.10}
    \setlength\tabcolsep{2.0mm}
    \caption{Quantitative comparison of different methods on PASCAL-Context dataset. * denotes the reproduced performance of methods based on the ViT-large backbone in \cite{ye2023taskexpert}.
    } \label{tab:pascal_context_comp} 
    \begin{tabular}{c|c|c|ccccc|cc} \hline 
    \bd{Method}  &  \bd{Publication} & \bd{Backbone} & \thead{\bd{Semseg} \\ mIoU $\uparrow$ } & \thead{\bd{Parsing} \\ mIoU $\uparrow$ } & \thead{\bd{Saliency} \\ maxF $\uparrow$ } 
    & \thead{\bd{Normal} \\ mErr $\downarrow$ } & \thead{\bd{Boundary} \\ odsF $\uparrow$ }  & \thead{\bd{FLOPs} \\ (G)} & \thead{\bd{\#Param} \\ (M) } \\ 
       \hline 
       PAD-Net \cite{xu2018pad}             & CVPR'18 & HRNet18   &  53.60 & 59.60   & 65.80  & 15.30   & 72.50  & 124  &  81  \\ 
       MTI-Net \cite{vandenhende2020mti}    & ECCV'20 & HRNet18   &  61.70 & 60.18   & 84.78  & 14.23   & 70.80  & 161  & 128  \\  
       ATRC \cite{bruggemann2021exploring}  & ICCV'21 & HRNet18   &  67.67 & 62.93   & 82.29  & 14.24   & 72.42  & 216  & 96   \\ \hline
       PAD-Net* \cite{xu2018pad}            & CVPR'18  & ViT-large &  78.01 & 67.12   & 79.21  & 14.37   & 72.60  & 773  & 330  \\ 
       MTI-Net* \cite{vandenhende2020mti}   & ECCV'20  & ViT-large &  78.31 & 67.40   & 84.75  & 14.67   & 73.00  & 774  & 851  \\  
       ATRC* \cite{bruggemann2021exploring} & ICCV'21  & ViT-large &  77.11 & 66.84   & 81.20  & 14.23   & 72.10  & 871  & 340  \\
       InvPT \cite{ye2022inverted}           & ECCV'22  & ViT-large &  79.03 & 67.61   & 84.81  & 14.15   & 73.00  & 669  & 423  \\
       TaskPrompter \cite{ye2022taskprompter} & ICLR'23 & ViT-large &  80.89 & 68.89   & 84.83  & 13.72   & 73.50  & 497  & 401  \\
       TaskExpert \cite{ye2023taskexpert}    & ICCV'23  & ViT-large &  80.64 & 69.42   & 84.87  & 13.56   & 73.30  & 622  & 420  \\ \hline
       \bd{Ours}                            & -  & ViT-large & \bd{81.41} & \bd{70.52}  
                                            & \bd{84.90} & \bd{13.51}  &\bd{75.42} &  571  & 407 \\ \hline 
    \end{tabular}  
\end{table*}

\myParaP{Task-specific router network.} 
The router network is important for generating task-specific 
gates, which decide how to activate the experts and 
assemble their features.
We ablate several settings of the router network and the results are shown in Tab.~\ref{tab:ablation_router}.
Adding the position-aware branch to the basic router network can improve 
the MTL gain by $+$0.17.
The position-aware branch can obtain more context information, which 
benefits for the router network.
Moreover, when replacing the router's input from the learnable parameters 
to the sample features, the MTL gain increases by $+$0.16, 
which demonstrates that dynamic information from samples 
is vital for gates.

\subsection{Comparisons with Other Methods} 
The quantitative comparisons with previous state-of-the-art (SOTA) methods are shown in \tabref{tab:pascal_context_comp} and \tabref{tab:nyud_comp}.
It can be seen that our method clearly outperforms previous methods 
in terms of all metrics on both the PASCAL-Context dataset and the NYUDv2 dataset.
In particular, on the PASCAL-Context dataset, for semantic segmentation, 
human parsing, and boundary detection, the performance of our method 
outperforms the previous best method by $+$0.52 mIoU, $+$1.10 mIoU and $+$1.92 odsF, respectively.

\begin{figure}[t]
  \centering
  \small
  \setlength\tabcolsep{1pt}
  \setlength{\abovecaptionskip}{5pt}
  \begin{overpic}[width=\linewidth]{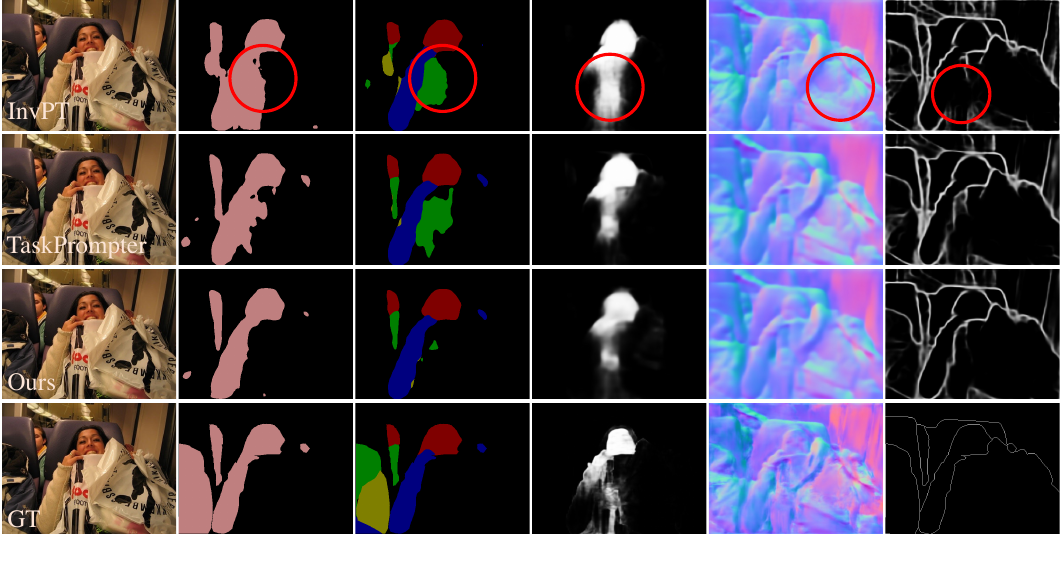}
  \put(3.0, 1.0){Image}
  \put(18.0, 1.0){Semseg}
  \put(35.0, 1.0){Parsing}
  \put(51.0, 1.0){Saliency}
  \put(68.5, 1.0){Normal}
  \put(83.5, 1.0){Boundary}
  \end{overpic}
  \caption{Qualitative comparison among different methods, including InvPT \cite{ye2022inverted}, 
  TaskPrompter \cite{ye2022taskprompter}, and ours. 
  Best viewed with zoom-in. It can be seen that our method achieves better visual results than other methods on all five tasks 
  thanks to the proposed MLoRE module.
  }\label{fig:vis_sota}

\end{figure}
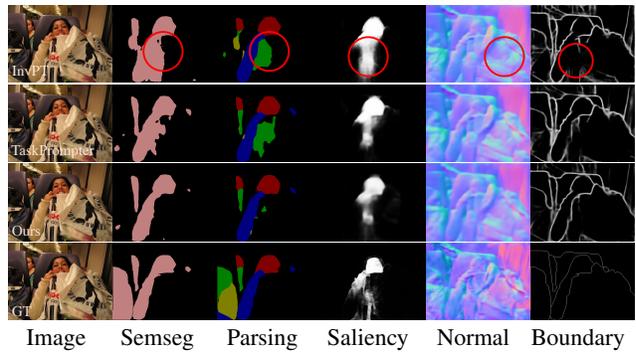

Previous methods, $\mbox{M}^3\mbox{ViT}$ \cite{fan2022m3vit}, Mod-Squad \cite{chen2023mod}, 
and TaskExpert \cite{ye2023taskexpert} all utilize the MoE technique in their network.
However, our method shows performance superior to theirs, which demonstrates the effectiveness 
of the MLoRE module.
Compared with the decoder-focused method, TaskExpert, the performance of semantic segmentation, 
human parsing, and object boundary is significantly improved by $+$0.77 mIoU, $+$1.10 mIoU 
and $+$2.12 odsF, but using fewer parameters and FLOPs.
Furthermore, we also present an intuitive visualization comparison among different methods in \figref{fig:vis_sota}.
Our method can generate better visualization results than the previous 
SoTA methods, especially for semantic segmentation, human parsing 
, and object boundary detection.
More visualization comparisons \yyq{are} in the supplemental material.

\begin{table}[t]
    \small
    \centering
    \caption{Quantitative comparison of different methods on the NYUD-v2 dataset. Our method performs the best on all four tasks.}
    \label{tab:nyud_comp}
    \tablestyle{2.4pt}{1.10}
    \vspace{-5pt}
    \begin{tabular}{c|c|cccc} \hline 
    \bd{Method} & \bd{Backbone} & \thead{\bd{Semseg} \\ mIoU $\uparrow$ } & \thead{\bd{Depth} \\ RMSE $\downarrow$ } 
    & \thead{\bd{Normal} \\ mErr $\downarrow$ } & \thead{\bd{Boundary} \\ odsF $\uparrow$ } \\ 
       \hline 
       PAD-Net \cite{xu2018pad}              &  HRNet18  &   36.61  & 0.6246 & 20.88  & 76.38 \\ 
       MTI-Net \cite{vandenhende2020mti}     &  HRNet48  &   45.97  & 0.5365 & 20.27  & 77.86 \\  
       ATRC \cite{bruggemann2021exploring}   &  HRNet48  &   46.33  & 0.5363 & 20.18  & 77.94 \\
       InvPT \cite{ye2022inverted}           & ViT-large &   53.56  & 0.5183 & 19.04  & 78.10 \\
       TaskPrompter \cite{ye2022taskprompter}& ViT-large &   55.30  & 0.5152 & 18.47  & 78.20 \\
       TaskExpert \cite{ye2023taskexpert}    & ViT-large &   55.35  & 0.5157 & 18.54  & 78.40 \\
       \bd{Ours}                           & ViT-large &\textbf{55.96}  & \textbf{0.5076} & \textbf{18.33}  & \textbf{78.43} \\
     \hline 
    \end{tabular}  
\end{table}

\subsection{Efficient MTL Models} 
We also apply our MLoRE module to the ViT-small backbone 
to check the performance of the efficient models.
Specifically, the channel number of the decoder decreases from 384 to 192.
As shown in \tabref{tab:moe_comp}, using about 35\% GFLOPs 
of TaskExpert, our method can achieve a highly competitive result.
In particular, the performance of the semantic segmentation 
and object boundary is improved by 0.6\% mIoU and 1.01\% odsF
while the metrics of other tasks are close to TaskExpert.
Furthermore, the parameters are fewer than TaskExpert by 11M.

\subsection{Conclusions}
We present a novel decoder-focused multi-task learning method, called MLoRE. We delve deep into the standard mixture-of-experts (MoE) technique and improve it for dense multi-task learning from two aspects.
First, to address the neglected global relationship 
modeling in the MoE, we add a simple generic convolution path to MoE,
enabling different task features to be able to share this path.
Furthermore, we apply the low-rank format of a vanilla convolution to different expert networks to free the MoE from high computational cost and a large number of parameters when increasing the number of experts.
Experiments demonstrate that the proposed method clearly outperforms 
previous state-of-the-art methods on all metrics.

\begin{table}[t!]
    \small
    \centering
    \caption{Quantitative comparison of the MoE-based efficient models 
    on PASCAL-Context dataset. }
    \label{tab:moe_comp} 
    \vspace{-5pt}
    \renewcommand{\arraystretch}{1.12}
    \setlength\tabcolsep{0.4mm} \vspace{2pt}
    \begin{tabular}{c|ccccc|cc}  \hline 
    \bd{Method}  & \thead{\bd{Semseg} \\ mIoU $\uparrow$ } & \thead{\bd{Parsing} \\ mIoU $\uparrow$ } & \thead{\bd{Sal.} \\ maxF $\uparrow$ } 
    & \thead{\bd{Nor.} \\ mErr $\downarrow$ } & \thead{\bd{Bound.} \\ odsF $\uparrow$ }  & \thead{\bd{FLOPs} \\ (G)} &  \thead{\bd{\#Param} \\ (M) } \\ 
       \hline 
        $\mbox{M}^3\mbox{ViT}$   &  72.80   & 62.10    &  66.30   &   14.50   & 71.70  & 420  & 42     \\ 
        Mod-Squad                &  74.10   &  62.70   &  66.90   &   13.70   & 72.00  & 420  & 52     \\
        TaskExpert               &  75.04   &  62.68   &  84.68   &   14.22   & 68.80  & 204  & 55     \\
        \bd{ours}                &  75.64   &  62.65   &  84.70   &   14.43   & 69.81  & 72   & 44     \\
    \hline 
    \end{tabular}  
    \vspace{-10pt}
\end{table}

\myPara{Acknowledgments.}
This research was supported by NSFC (NO. 62276145), the Fundamental Research Funds for the Central Universities (Nankai University, 070-63223049), CAST through
Young Elite Scientist Sponsorship Program (No. YESS20210377). Computations were supported by the Supercomputing Center of Nankai University (NKSC).

\newpage
\appendix
\section{More Implementation Details}
\subsection{MLoRE Module Stacking}
In our method, we stack two MLoRE modules after each selected backbone layer.
In the first MLoRE module, the lightweight task-specific $1\times1$ convolutions 
are utilized to project the backbone feature to different task features.
In the second MLoRE module, as the task-specific features have been split already, 
we utilize $1\times1$ convolutions directly to deal with task-specific features.
Moreover, as the MLoRE is a linear module, to introduce 
the non-linearity into the decoder, we insert the task-specific non-linear block 
between the two MLoRE modules and after the last MLoRE module.
Each non-linear block is composed of a BatchNorm-GELU-Linear structure.
When fusing the features from different scales, we generate a pixel-wise mask 
from the concatenated features and leverage it to weight the feature from different scales.

\subsection{MoE Optimization}
Following previous MoE-based MTL methods~\cite{fan2022m3vit, chen2023mod}, 
we utilize the noising gating and load-balancing loss proposed by 
Shazeer \etal \cite{shazeer2017outrageously}, 
which is a common practice in sparse MoE training~\cite{shazeer2017outrageously, riquelme2021scaling}.

One may concern that without the load-balancing loss, it will have a higher possibility 
for an expert to be activated by all the tasks on the same sample.
However, we can't discard the load-balancing loss to construct the global relationships 
across all the tasks in one expert for two reasons.
Firstly, discarding the load-balancing loss will weaken MoE's ability to dynamically choose different experts for different samples, which is opposite to our motivation for using MoE.
Secondly, without the load-balancing loss, most experts will be less or never activated, which 
will hurt the capacity of MoE.
On the contrary, the proposed task-sharing generic path will not harm the ability of dynamic routing 
and the capacity of MoE. 
We will prove the necessity of load-balancing loss in the following experiments.

In addition, our MLoRE is trained with the top-$k$ constraint.
When training MoE without top-$k$ constraint, we found each expert would be shared 
by all the tasks.
However, we empirically found that this might make the optimization process difficult 
and harm MoE's ability to build relationships in a subset of tasks.
As a result, although it can build global task relationships, the performance 
is highly influenced without the top-$k$ constraint, as described in the left figure 
of Fig. 5 in the main paper.
It can be seen that the performance of the setting without top-$k$ constraint 
is lower than the top-$k$ settings.
On the contrary, our proposed task-sharing generic path can explicitly build 
global task relationships while MoE still builds relationships in a subset of tasks.  

At last, we also enable the router to predict one extra scaling value to weight the feature 
from the task-specific low-rank expert path.
The scaling value doesn't get involved in top-$k$ selecting process.

\subsection{Re-parameterization During Training}
Since the re-parameterization can speed up the forward propagation, it is natural to 
ask if we can extend the re-parameterization to the training phase 
for better training efficiency.
However, in our MLoRE module, the re-parameterization can only be performed 
at the inference stage.
The re-parameterization in training will largely influence the training-time behaviour 
and the reason is the BatchNorm layer in our MLoRE module.
We follow RepVGG~\cite{ding2021repvgg} to set BatchNorm in our task-sharing low-rank expert path, which is important for the re-parameterization-based method as stated in Sec.~4.2 of RepVGG.
When a BatchNorm layer merges with a convolution layer in training, 
the feature statistics for this BatchNorm layer will be hard to perform.

\section{Additional Study on MLoRE Module}
\subsection{Number of MLoRE Module}
The number of the MLoRE module at one scale would 
also influence the performance. 
We conduct a series of experiments on it, and the results are shown in \tabref{tab:ablation_number}.
It can be seen that when increasing the number of the MLoRE modules from 1 to 2, 
the MTL gain is increased from -1.25 to -0.58.
When further stacking 1 MLoRE module, we do not observe obvious 
performance gain (-0.74 \emph{v.s.} -0.58).
Thus, in our paper, we set the number of the MLoRE module 
at each scale to 2.

\begin{table}[t]
    \small
    \centering
    \caption{Ablation on the number of MLoRE module 
    on PASCAL-Context dataset. }
    \label{tab:ablation_number}
    \renewcommand{\arraystretch}{1.1}
    \setlength\tabcolsep{0.7mm}
    \vspace{-10pt}
    \begin{tabular}{c|ccccc|c} \hline 
    \thead{\bd{Number}} & \thead{\bd{Semseg} \\ mIoU $\uparrow$ } & \thead{\bd{Parsing} \\ mIoU $\uparrow$ } & \thead{\bd{Saliency} \\ maxF $\uparrow$ } 
    & \thead{\bd{Normal} \\ mErr $\downarrow$ } & \thead{\bd{Boundary} \\ odsF $\uparrow$ }  & \thead{\bd{MTL} \\ $\Delta_m$ $\uparrow$} \\ 
       \hline 
        1   & 78.80 & 66.83 & 85.15 & 13.57 & 73.40 & -1.25 \\ 
        2   & 79.26 & 67.82 & 85.30 & 13.65 & 74.69 & -0.58 \\
        3   & 79.05 & 67.94 & 85.07 & 13.72 & 74.75 & -0.74 \\
    \hline 
    \end{tabular}  
\end{table}

\subsection{Low-rank Task-Sharing Generic Path}
We further ablate the effectiveness of the low-rank task-sharing generic path.
We utilize the low-rank format of a vanilla 3×3 convolution in the task-sharing generic path 
to explore whether we can design a lighter module with a fully low-rank structure in MLoRE.
The results are shown in \tabref{tab:ablation_gc_rank}.
With the increase of the rank, the performance improves on most of the tasks.
When using the vanilla 3x3 convolution, its performance outperforms the 
low-rank settings by a large margin.
This result shows that it is beneficial to use the vanilla 3x3 convolution rather 
than its low-rank format to construct the task-sharing generic path.

\begin{table}[t]
    \small
    \centering
    \caption{Ablation on the rank setting of the low-rank task-sharing generic path 
    on PASCAL-Context dataset. }
    \label{tab:ablation_gc_rank}
    \renewcommand{\arraystretch}{1.1}
    \setlength\tabcolsep{0.7mm}
    \vspace{-10pt}
    \begin{tabular}{c|ccccc|c} \hline 
    \thead{\bd{Rank}} & \thead{\bd{Semseg} \\ mIoU $\uparrow$ } & \thead{\bd{Parsing} \\ mIoU $\uparrow$ } & \thead{\bd{Saliency} \\ maxF $\uparrow$ } 
    & \thead{\bd{Normal} \\ mErr $\downarrow$ } & \thead{\bd{Boundary} \\ odsF $\uparrow$ }  & \thead{\bd{MTL} \\ $\Delta_m$ $\uparrow$} \\ 
       \hline 
        16          & 78.53 & 66.71 & 85.27 & 13.68 & 73.70 & -1.41 \\ 
        128         & 78.78 & 67.01 & 85.08 & 13.70 & 73.98 & -1.26 \\
        vanilla (3x3)   & 79.26 & 67.82 & 85.30 & 13.65 & 74.69 & -0.58 \\
    \hline
    \end{tabular}  
\end{table}

\subsection{Detailed Results for Number of Task-Sharing Low-Rank Experts and Top-k Selection}
We present the detailed performance of the number of task-sharing low-rank experts and top-k selection for every task in \tabref{tab:ablation_expert_num} and \tabref{tab:ablation_topk}.
It can be seen that increasing the number of experts can achieve better performance 
on most of the tasks, which is also verified in previous work~\cite{chen2023mod}.
This also proves the necessity of introducing the linear and low-rank structure 
into the MoE.

\begin{table}[t]
    \small
    \centering
    \caption{Ablation on the number of experts 
    on PASCAL-Context dataset. }
    \label{tab:ablation_expert_num}
    \renewcommand{\arraystretch}{1.1}
    \setlength\tabcolsep{0.7mm}
    \vspace{-10pt}
    \begin{tabular}{c|ccccc|c} \hline
    \thead{\bd{Expert} \\ \bd{Number}} & \thead{\bd{Semseg} \\ mIoU $\uparrow$ } & \thead{\bd{Parsing} \\ mIoU $\uparrow$ } & \thead{\bd{Saliency} \\ maxF $\uparrow$ } 
    & \thead{\bd{Normal} \\ mErr $\downarrow$ } & \thead{\bd{Boundary} \\ odsF $\uparrow$ }  & \thead{\bd{MTL} \\ $\Delta_m$ $\uparrow$} \\ 
       \hline 
        0          & 78.30 & 67.43 & 85.19 & 14.03 & 74.52 & -1.58 \\ 
        5          & 78.73 & 67.47 & 85.16 & 13.76 & 74.36 & -1.11 \\
        10         & 78.95 & 67.75 & 85.24 & 13.63 & 74.51 & -0.72 \\
        15         & 79.26 & 67.82 & 85.30 & 13.65 & 74.69 & -0.58 \\
        20         & 79.19 & 67.92 & 85.13 & 13.57 & 74.55 & -0.53 \\
    \hline 
    \end{tabular}  
\end{table}

\begin{table}[t]
    \small
    \centering
    \caption{Ablation of the top-$k$ with 15 experts
    on PASCAL-Context dataset. }
    \label{tab:ablation_topk}
    \renewcommand{\arraystretch}{1.1}
    \setlength\tabcolsep{0.7mm}
    \vspace{-10pt}
    \begin{tabular}{c|ccccc|c} \hline 
    \thead{\bd{Top-$k$}} & \thead{\bd{Semseg} \\ mIoU $\uparrow$ } & \thead{\bd{Parsing} \\ mIoU $\uparrow$ } & \thead{\bd{Saliency} \\ maxF $\uparrow$ } 
    & \thead{\bd{Normal} \\ mErr $\downarrow$ } & \thead{\bd{Boundary} \\ odsF $\uparrow$ }  & \thead{\bd{MTL} \\ $\Delta_m$ $\uparrow$} \\ 
       \hline 
        $k$=3          & 79.22 & 67.84 & 85.18 & 13.70 & 74.25 & -0.72 \\ 
        $k$=6          & 79.19 & 67.81 & 85.25 & 13.66 & 74.64 & -0.65 \\
        $k$=9          & 79.26 & 67.82 & 85.30 & 13.65 & 74.69 & -0.58 \\
        $k$=12         & 79.05 & 67.75 & 85.23 & 13.58 & 74.43 & -0.64 \\
        $k$=15         & 78.91 & 67.51 & 85.18 & 13.54 & 74.22 & -0.75 \\
    \hline
    \end{tabular}  
\end{table}

\begin{figure}[t]
    \centering
    \setlength{\abovecaptionskip}{4pt}
    \setlength\tabcolsep{1pt}
    \begin{overpic}[width=0.48\textwidth]{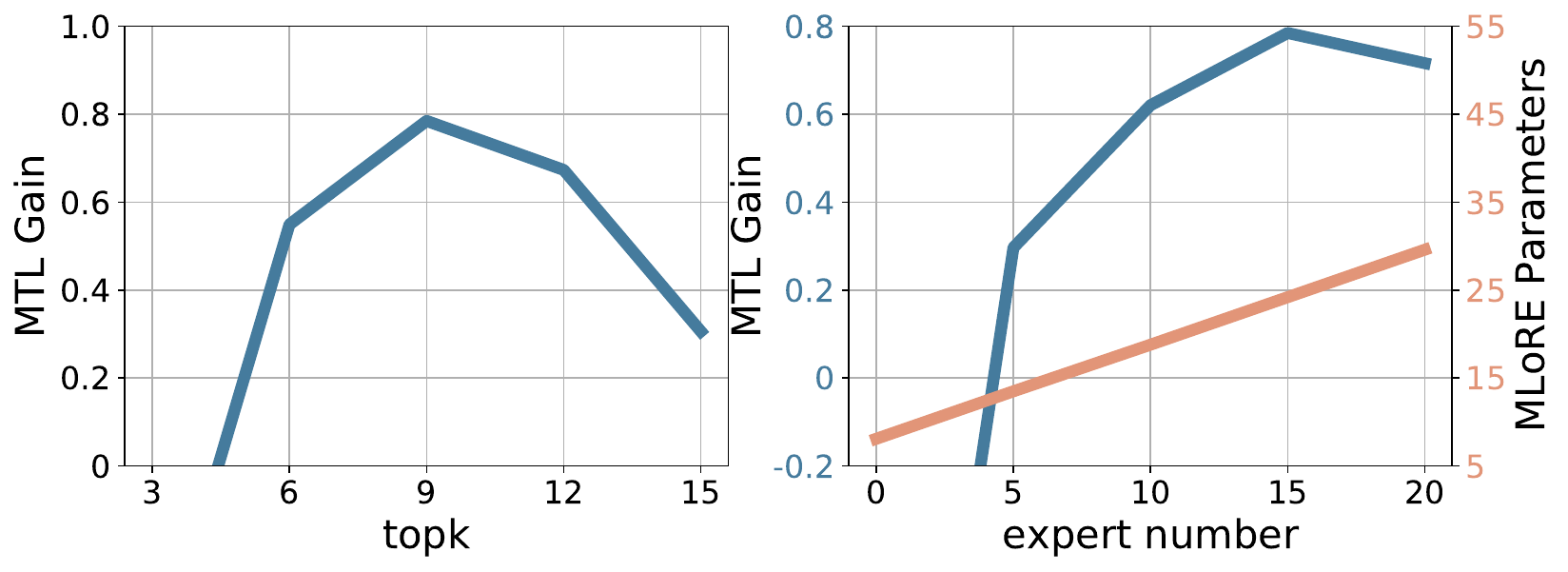}
    \end{overpic}
    \caption{Ablation study on the number of experts $N$ and the number 
    of activated experts $K$ on dataset NYUD-v2. In the right figure, we also present the parameter change 
    of the MLoRE module with the increase in the number of experts.
    }\label{fig:ablation_n_k_nyud}
\end{figure} 

\subsection{Effectiveness of Load-Balancing Loss}
We conduct extensive experiments to verify the effectiveness of load-balancing loss.
To prove its necessity for the MoE structure, we test three settings with 
different designs.
The results are shown in \tabref{tab:ablation_lbloss}.
It can be clearly seen that with the load-balancing loss, it will achieve better 
performance on most tasks in all three settings.
The MTL gain is also improved with the load-balancing loss.
The quantitative results demonstrate the effectiveness of the load-balancing loss 
for the MoE structure and motivates us to propose the task-sharing generic path 
rather than discard the load-balancing loss to build global relationships.

\begin{table}[t]
    \small
    \centering
    \caption{Ablation of the effectiveness of load-balancing loss in different settings 
    on PASCAL-Context dataset. 
    \bd{MoE}: baseline with the standard MoE structure. 
    \bd{LoRE}: baseline with the task-sharing low-rank expert path.
    \bd{Ours}: our MLoRE is equipped with all the components.
    \bd{w/o LB loss}: without the load-balancing loss.}
    \label{tab:ablation_lbloss}
    \renewcommand{\arraystretch}{1.1}
    \setlength\tabcolsep{0.7mm}
    \vspace{-10pt}
    \begin{tabular}{c|ccccc|c} \hline
    \thead{\bd{Settings}} & \thead{\bd{Semseg} \\ mIoU $\uparrow$ } & \thead{\bd{Parsing} \\ mIoU $\uparrow$ } & \thead{\bd{Saliency} \\ maxF $\uparrow$ } 
    & \thead{\bd{Normal} \\ mErr $\downarrow$ } & \thead{\bd{Boundary} \\ odsF $\uparrow$ }  & \thead{\bd{MTL} \\ $\Delta_m$ $\uparrow$} \\ 
       \hline 
        MoE          & 78.56 & 66.78 & 85.17 & 13.58 & 73.91 & -1.20 \\ 
        w/o LB loss  & 78.32 & 66.41 & 85.14 & 13.58 & 73.81 & -1.40 \\
        \hline
        LoRE         & 78.38 & 66.21 & 85.15 & 13.71 & 73.53 & -1.71 \\
        w/o LB loss  & 78.04 & 66.05 & 85.10 & 13.69 & 73.60 & -1.80 \\
        \hline
        Ours         & 79.26 & 67.82 & 85.30 & 13.65 & 74.69 & -0.58 \\
        w/o LB loss  & 79.01 & 68.03 & 85.24 & 13.66 & 74.38 & -0.70 \\
    \hline 
    \end{tabular}  
\end{table}

\subsection{Ablation on NYUD-v2}
\begin{figure*} [h]
  \begin{subfigure}{.595\textwidth}
    \centering
    \includegraphics[width=1\linewidth]{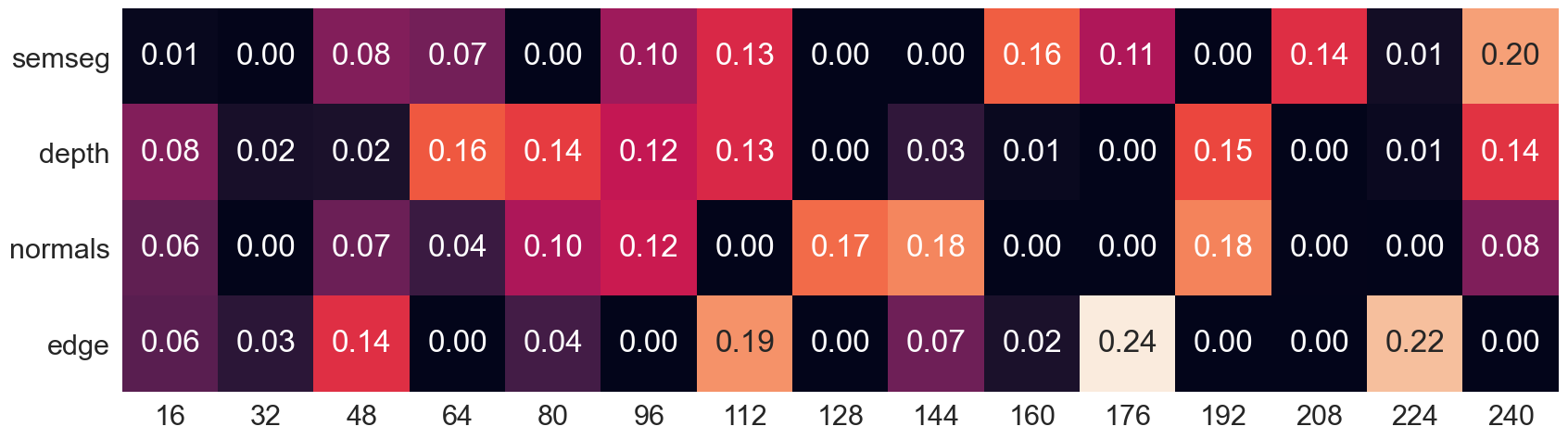}
\end{subfigure}
  \begin{subfigure}{.395\textwidth}
    \centering
    \includegraphics[width=1\linewidth]{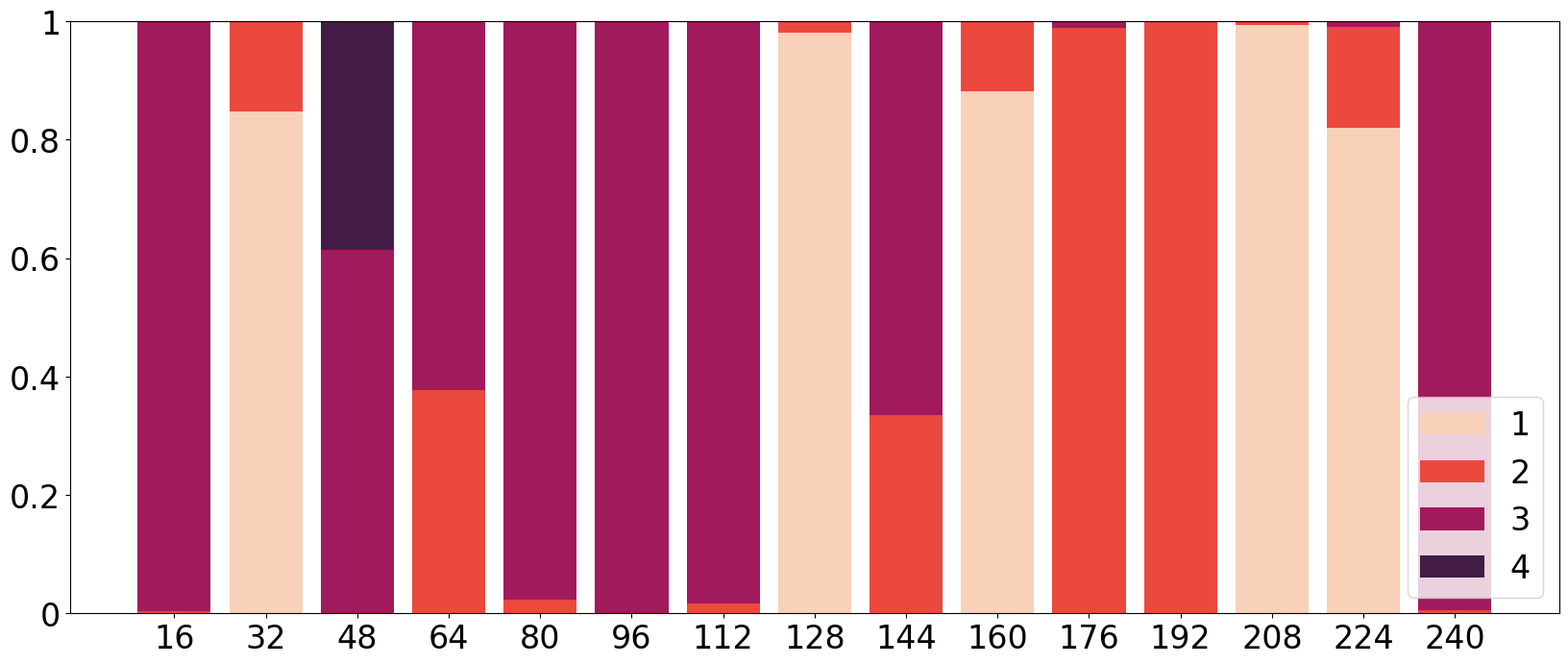}
\end{subfigure}
  \vspace{-5pt}
\caption{
    (a) The relations between tasks and low-rank experts on dataset NYUD-v2. 
    (b) The ratio of an expert activated by different numbers of tasks in the MLoRE module without the task-sharing generic path on dataset NYUD-v2.
    We can see that without the task-sharing generic path, there are only a few experts can be activated by all four tasks.
    Horizontal coordinates represent the ranks of different experts.
  }\label{fig:vis_relation_nyud}
  \vspace{-10pt}
\end{figure*}

We also conduct some important ablations on the NYUD-v2 dataset.
Specifically, we conduct the ablation on the number of task-sharing low-rank experts 
and top-k selection on NYUD-v2.
The results are shown in \figref{fig:ablation_n_k_nyud}.
In addition, we also show the relation visualization on NYUD-v2 in \figref{fig:vis_relation_nyud}
It can be seen that the results of these ablations still support the conclusions 
in our main paper.

\begin{table}[t]
    \small
    \centering
    \caption{Performance comparison between our method and the baseline based on ViT-L 
    on the NYUD-v2 and PASCAL-Context datasets. }
    \label{tab:baseline}
    \scriptsize
    \vspace{-8pt}
    \renewcommand{\arraystretch}{1.1}
    \setlength\tabcolsep{0.6mm} \vspace{4pt}
    \begin{tabular}{c|ccccc|cccc}   \hline
    ~  & \multicolumn{5}{c|}{\bd{Pascal-Context}} & \multicolumn{4}{c}{\bd{NYUDv2}} \\ 
    \hline
    \bd{Method}  & \bd{Semseg} & \bd{Parsing} & \bd{Sal.} & \bd{Nor.}  & \bd{Bound.} & \bd{Semseg} & \bd{Depth} & \bd{Nor.}  & \bd{Bound.} \\
       \hline 
        Baseline  &  78.59   &  67.78   &  84.43   &   13.87   & 70.26 & 54.55 & 0.6043 & 18.62 & 75.21 \\
        Ours  & 81.41 & 70.52 & 84.90 & 13.51 & 75.42 & 55.96 & 0.5076 & 18.33 & 78.43 \\
    \hline 
    \end{tabular}  
\end{table}

\subsection{More Quantitative Results}
We show the performance of the baseline models based on ViT-L in \tabref{tab:baseline}.
Furthermore, we conduct experiments to evaluate the performance 
based on HRNet18~\cite{sun2019deep}.
The results are shown in \tabref{tab:hrnet}. 
Our method outperforms other methods in terms of $\Delta_m$ by a large margin.

\begin{table}[t]
    \small
    \centering
    \caption{Ablation of the top-$k$ with 15 experts
    on PASCAL-Context dataset. }
    \label{tab:hrnet}
    \renewcommand{\arraystretch}{1.1}
    \setlength\tabcolsep{0.7mm}
    \vspace{-10pt}
    \begin{tabular}{c|ccccc|c} \hline 
    \thead{\bd{Method} \\ (HRNet18)} & \thead{\bd{Semseg} \\ mIoU $\uparrow$ } & \thead{\bd{Parsing} \\ mIoU $\uparrow$ } & \thead{\bd{Saliency} \\ maxF $\uparrow$ } 
    & \thead{\bd{Normal} \\ mErr $\downarrow$ } & \thead{\bd{Boundary} \\ odsF $\uparrow$ }  & \thead{\bd{MTL} \\ $\Delta_m$ $\uparrow$} \\ 
       \hline 
       MTI-Net & 61.70  &  60.18   &  84.78  &  14.23   &  70.80 & -2.12 \\
       ATRC & 62.69  &  59.42   &  84.70  &  14.20   &  70.96 & -1.98 \\
       Ours  &  62.43   &  60.78   &  84.85  &  14.05   &  71.84 & -1.13 \\
    \hline
    \end{tabular}  
\end{table}

\section{More Visual Results}

\subsection{More Visual Comparison Results}
\begin{figure*}[t]
  \centering
  \small
  \setlength\tabcolsep{1pt}
  \setlength{\abovecaptionskip}{5pt}
  \begin{overpic}[width=0.8\linewidth]{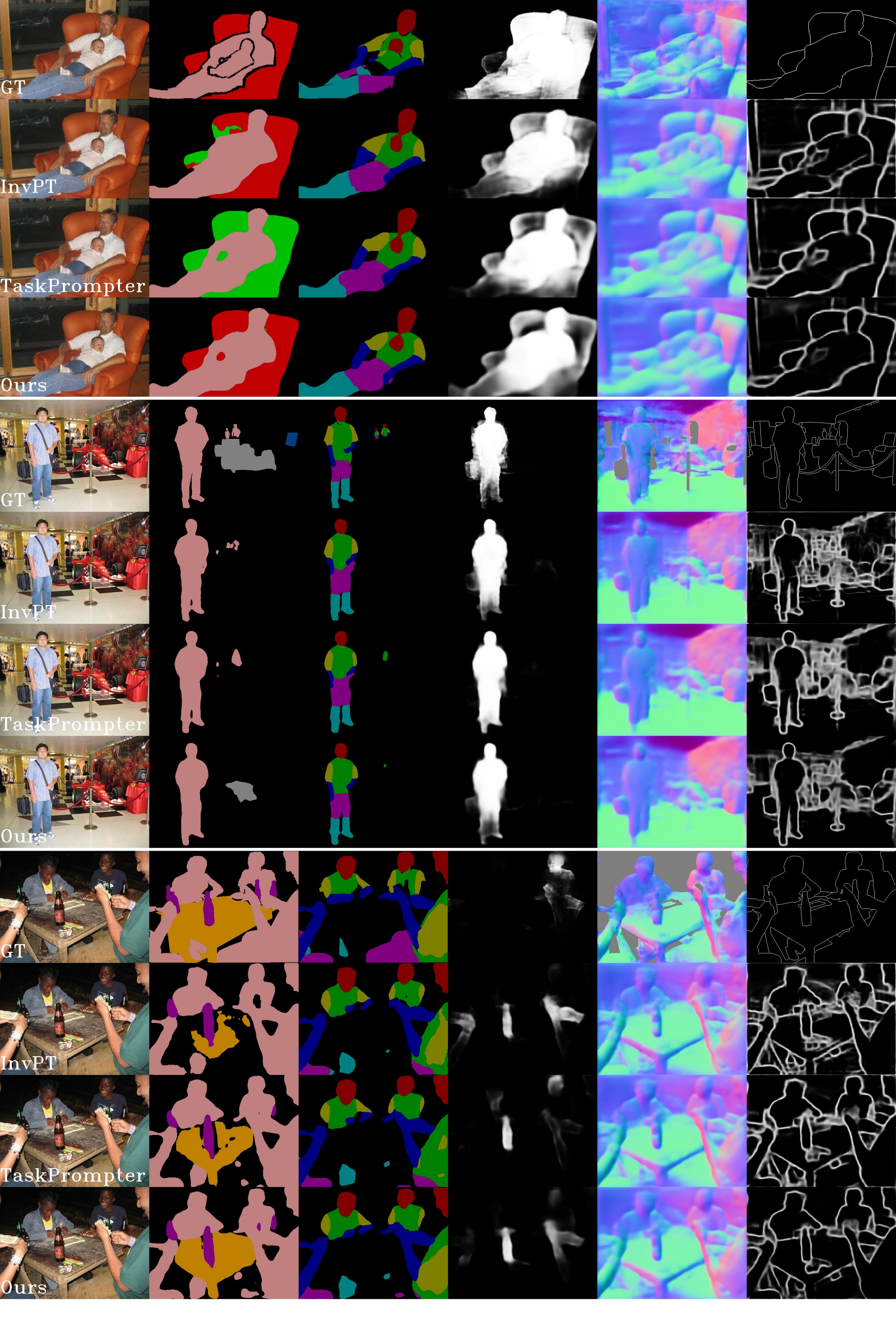}
  \put(4.0, 0.5){Image}
  \put(14.5, 0.5){Semseg}
  \put(26.0, 0.5){Parsing}
  \put(36.5, 0.5){Saliency}
  \put(48, 0.5){Normal}
  \put(58.5, 0.5){Boundary}
  \end{overpic}
  \caption{More Qualitative comparison on PASCAL-Context dataset among different methods, including InvPT \cite{ye2022inverted}, 
  TaskPrompter \cite{ye2022taskprompter}, and ours. 
  Best viewed with zoom-in. It can be seen that our method achieves better visual results than other methods on all five tasks thanks to the proposed MLoRE module.
  }\label{fig:vis_sota_pascal}

\end{figure*}

\begin{figure*}[t]
  \centering
  \small
  \setlength\tabcolsep{1pt}
  \setlength{\abovecaptionskip}{5pt}
  \begin{overpic}[width=0.60\linewidth]{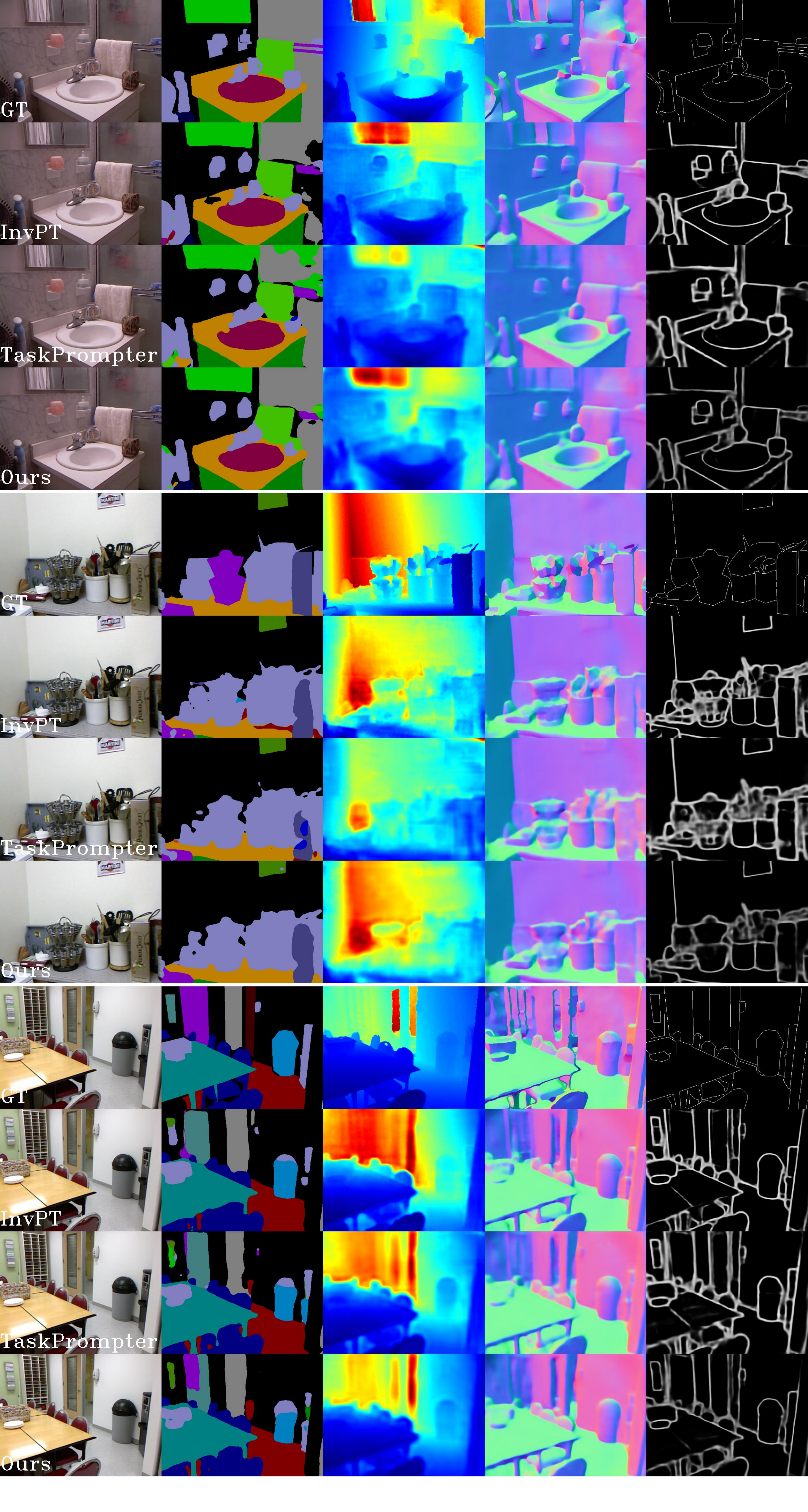}
  \put(3.0, 0.5){Image}
  \put(13.5, 0.5){Semseg}
  \put(25.0, 0.5){Depth}
  \put(35.5, 0.5){Normal}
  \put(45.5, 0.5){Boundary}
  \end{overpic}
  \caption{More Qualitative comparison on NYUD-v2 dataset among different methods, including InvPT \cite{ye2022inverted}, 
  TaskPrompter \cite{ye2022taskprompter}, and ours. 
  Best viewed with zoom-in. It can be seen that our method achieves better visual results than other methods on all five tasks thanks to the proposed MLoRE module.
  }\label{fig:vis_sota_nyud}

\end{figure*}

We present more qualitative results compared with the former SOTA methods, TaskPrompter~\cite{ye2022taskprompter} and InvPT~\cite{ye2022inverted}.
In \figref{fig:vis_sota_pascal} and \figref{fig:vis_sota_nyud}, we can see that 
our method generates better visual results than previous SOTA methods 
on most of the tasks.

\subsection{Relation Visualizations from More Layers}
\begin{figure*} [t]
  \begin{subfigure}{.49\textwidth}
    \centering
    \includegraphics[width=1\linewidth]{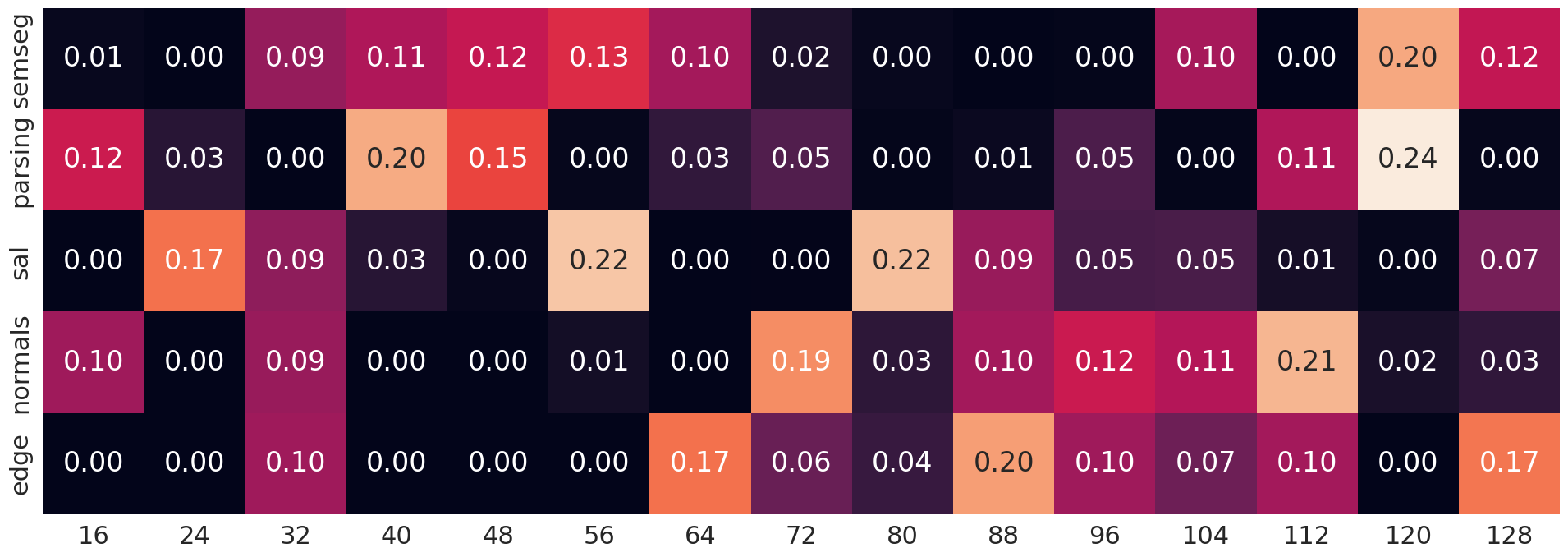}
\end{subfigure}
\begin{subfigure}{.49\textwidth}
    \centering
    \includegraphics[width=1\linewidth]{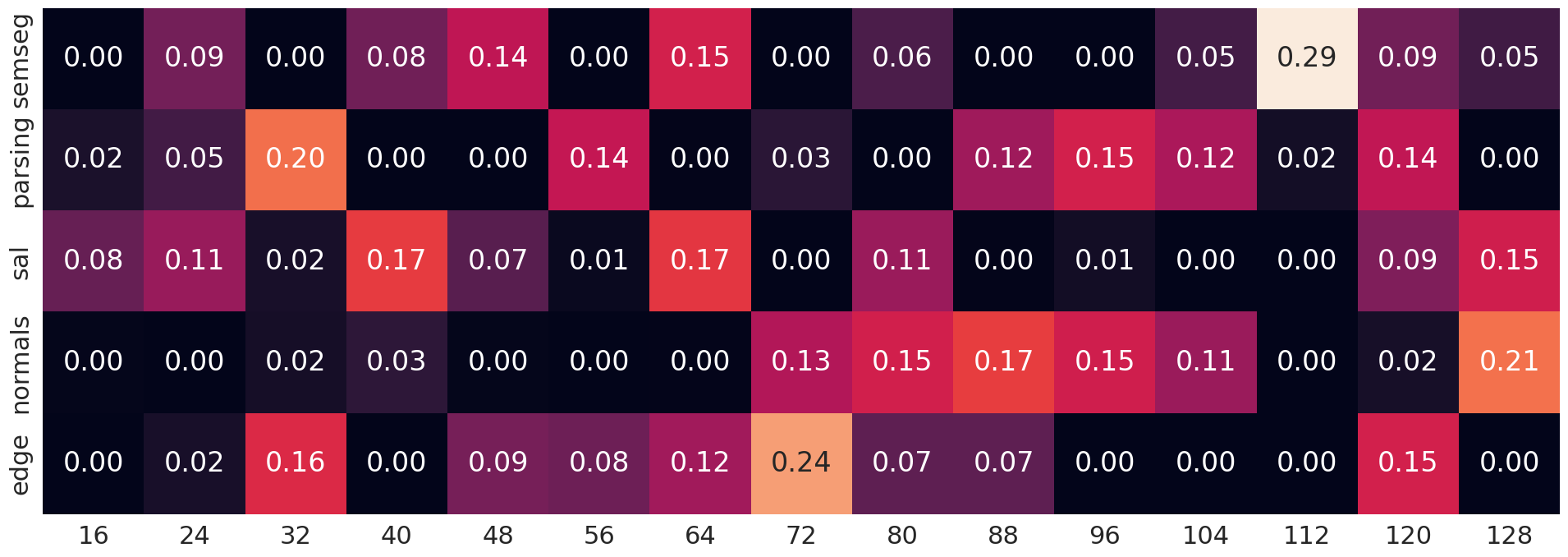}
\end{subfigure}
\begin{subfigure}{.49\textwidth}
    \centering
    \includegraphics[width=1\linewidth]{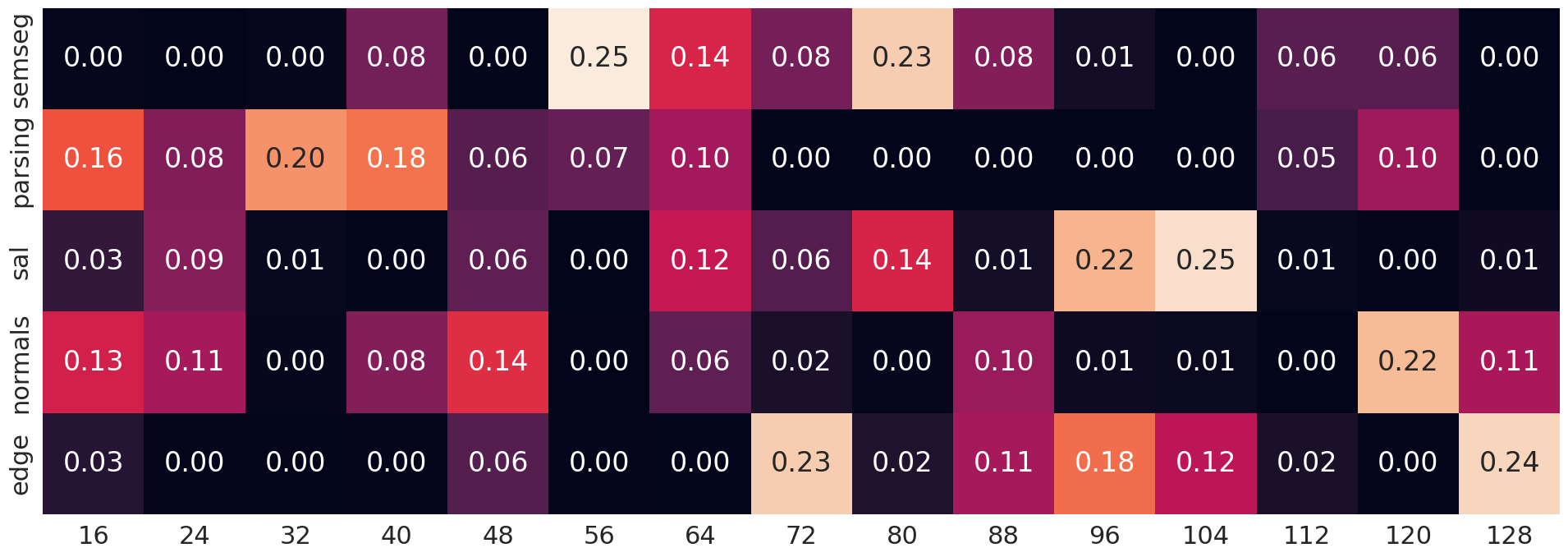}
\end{subfigure}
\begin{subfigure}{.49\textwidth}
    \centering
    \includegraphics[width=1\linewidth]{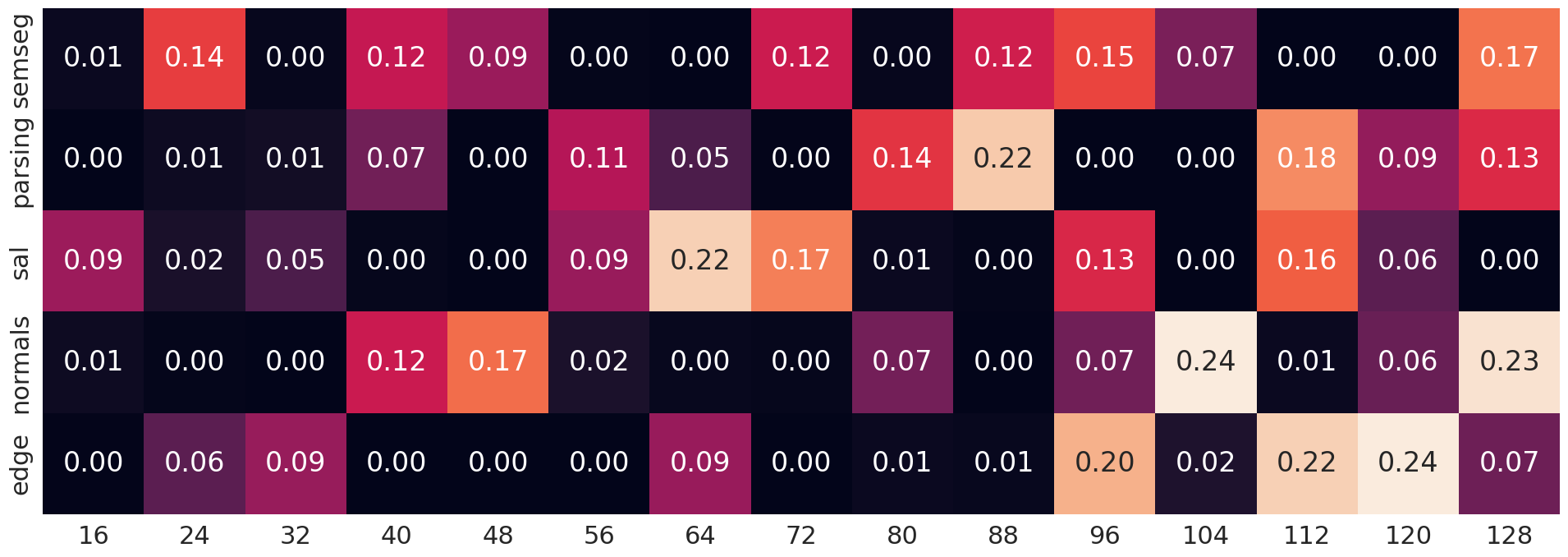}
\end{subfigure}
\begin{subfigure}{.49\textwidth}
    \centering
    \includegraphics[width=1\linewidth]{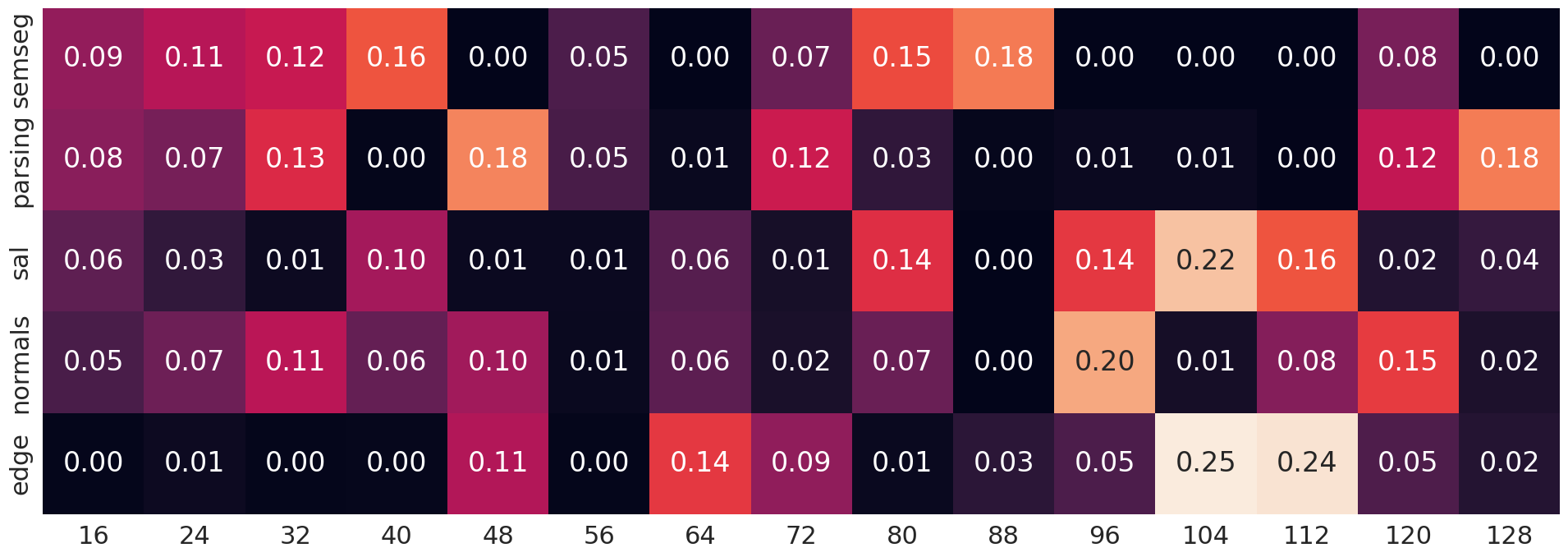}
\end{subfigure}
\begin{subfigure}{.49\textwidth}
    \centering
    \includegraphics[width=1\linewidth]{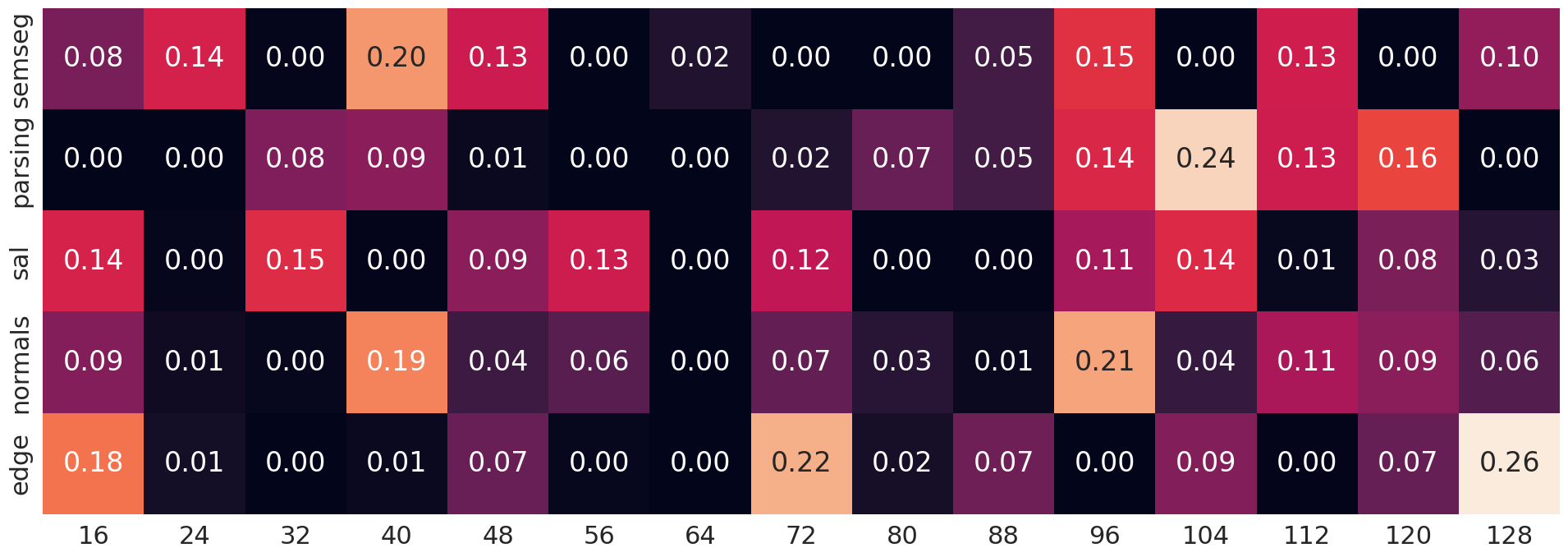}
\end{subfigure}
\begin{subfigure}{.49\textwidth}
    \centering
    \includegraphics[width=1\linewidth]{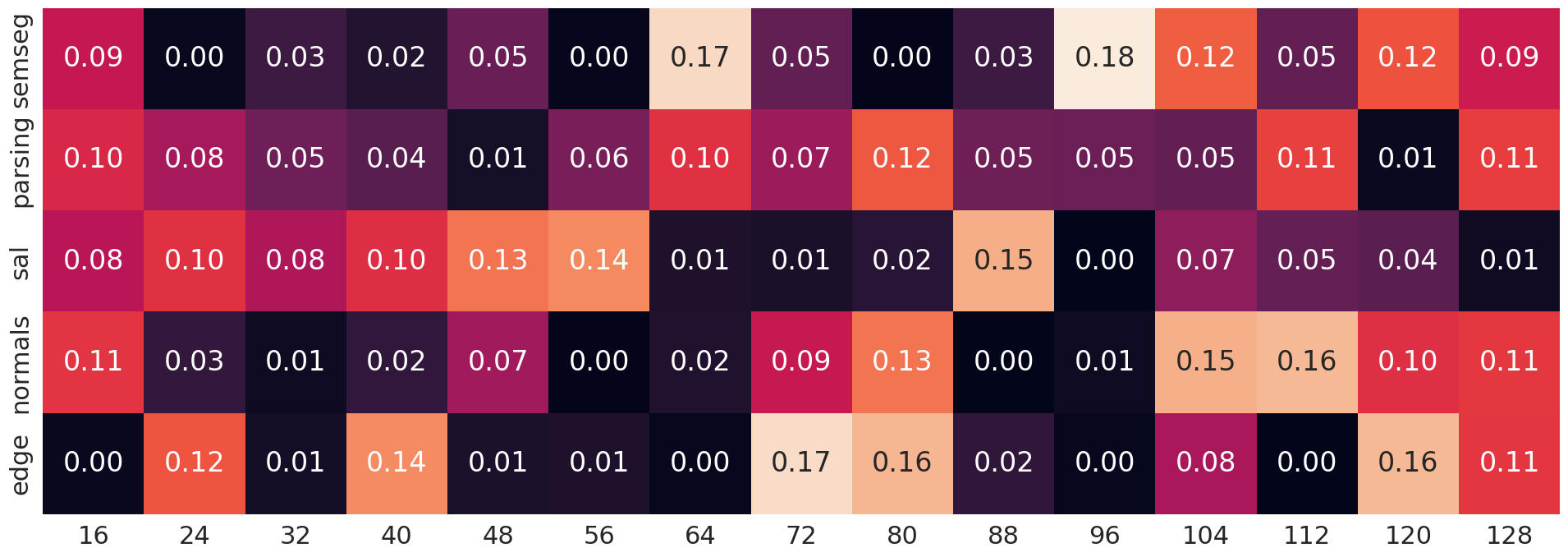}
\end{subfigure}
\begin{subfigure}{.49\textwidth}
    \centering
    \includegraphics[width=1\linewidth]{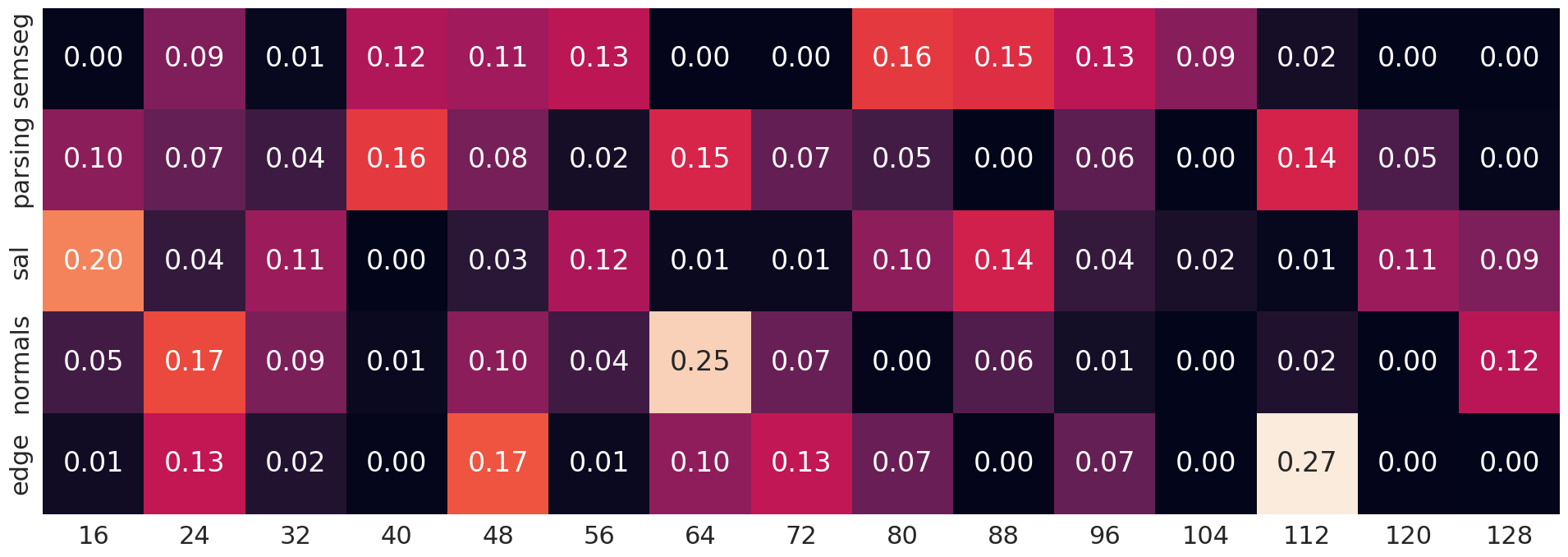}
\end{subfigure}
  \vspace{-5pt}
\caption{
    The relations between tasks and low-rank experts in all the MLoRE modules in our model.
    }\label{fig:vis_relationa_other}
  \vspace{-5pt}
\end{figure*}

\begin{figure*} [t]
  \begin{subfigure}{.49\textwidth}
    \centering
    \includegraphics[width=1\linewidth]{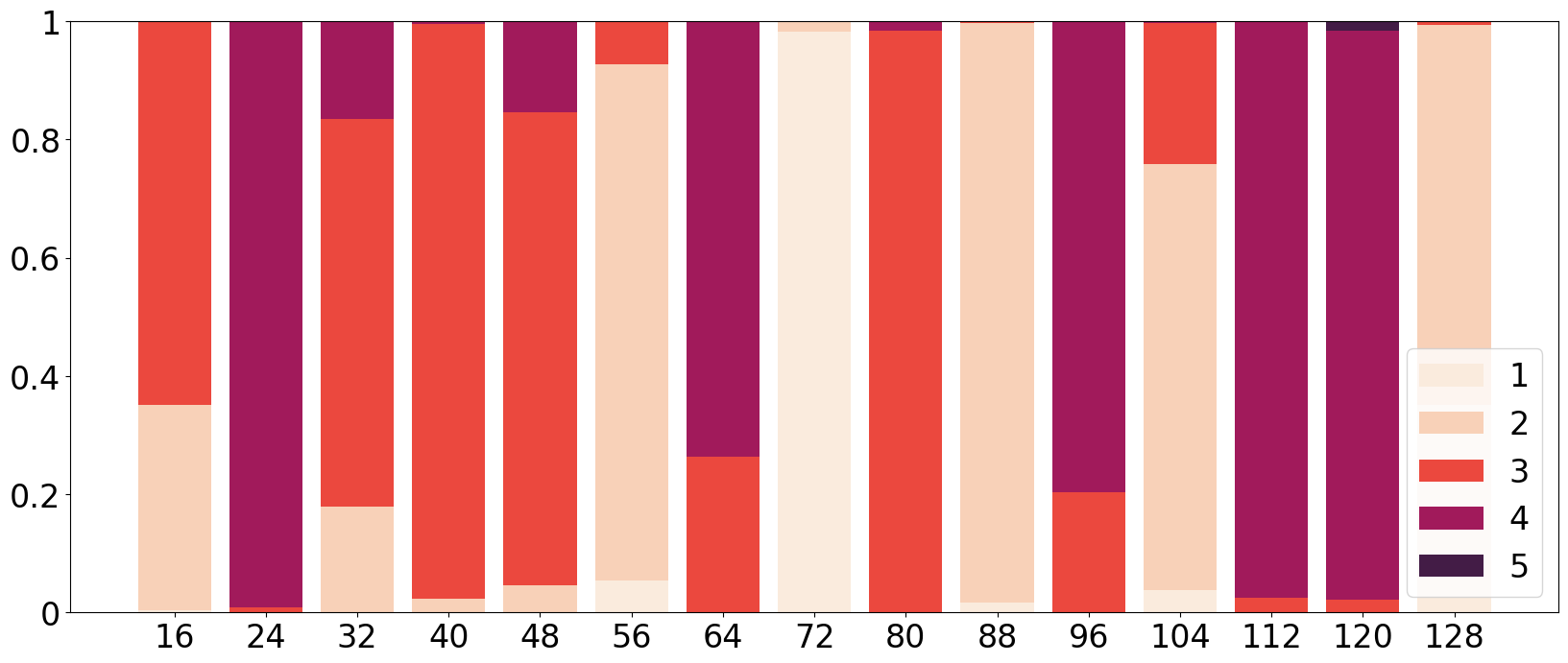}
\end{subfigure}
  \begin{subfigure}{.49\textwidth}
    \centering
    \includegraphics[width=1\linewidth]{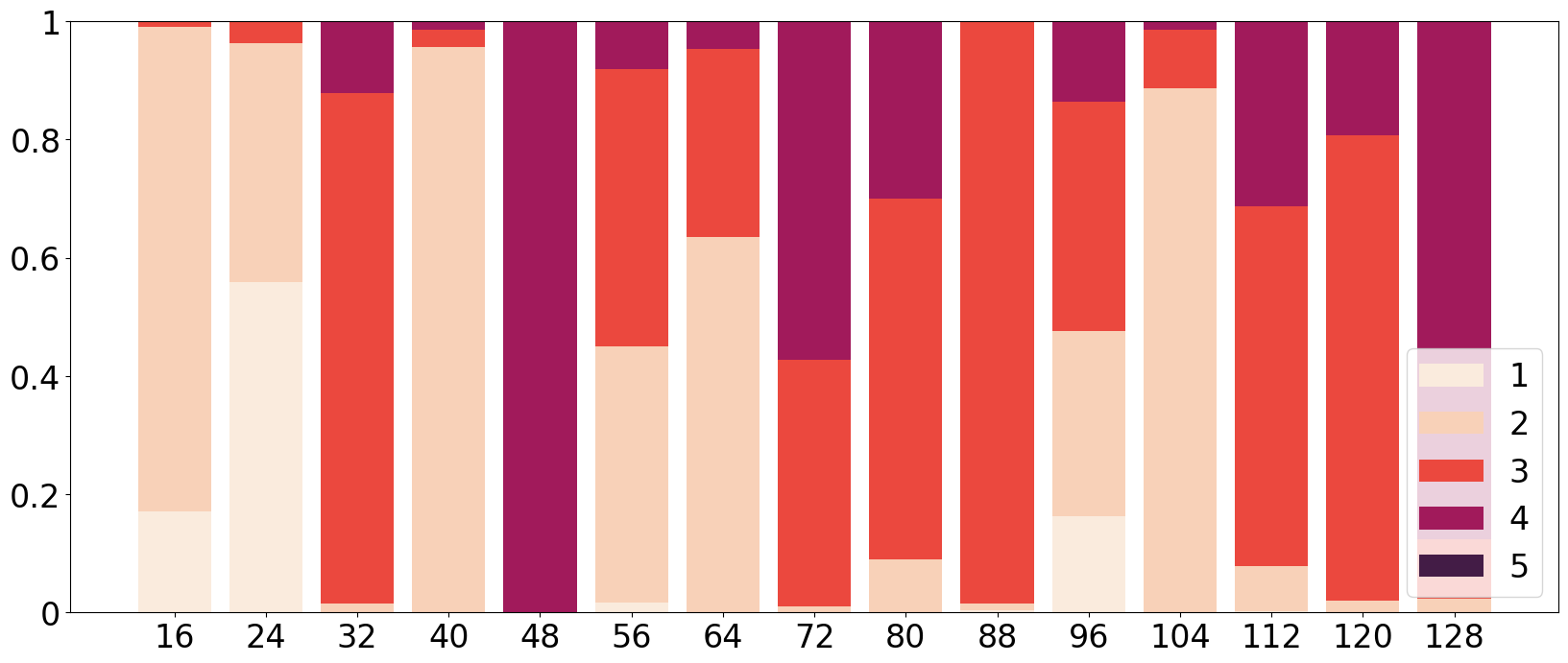}
\end{subfigure}
\begin{subfigure}{.49\textwidth}
    \centering
    \includegraphics[width=1\linewidth]{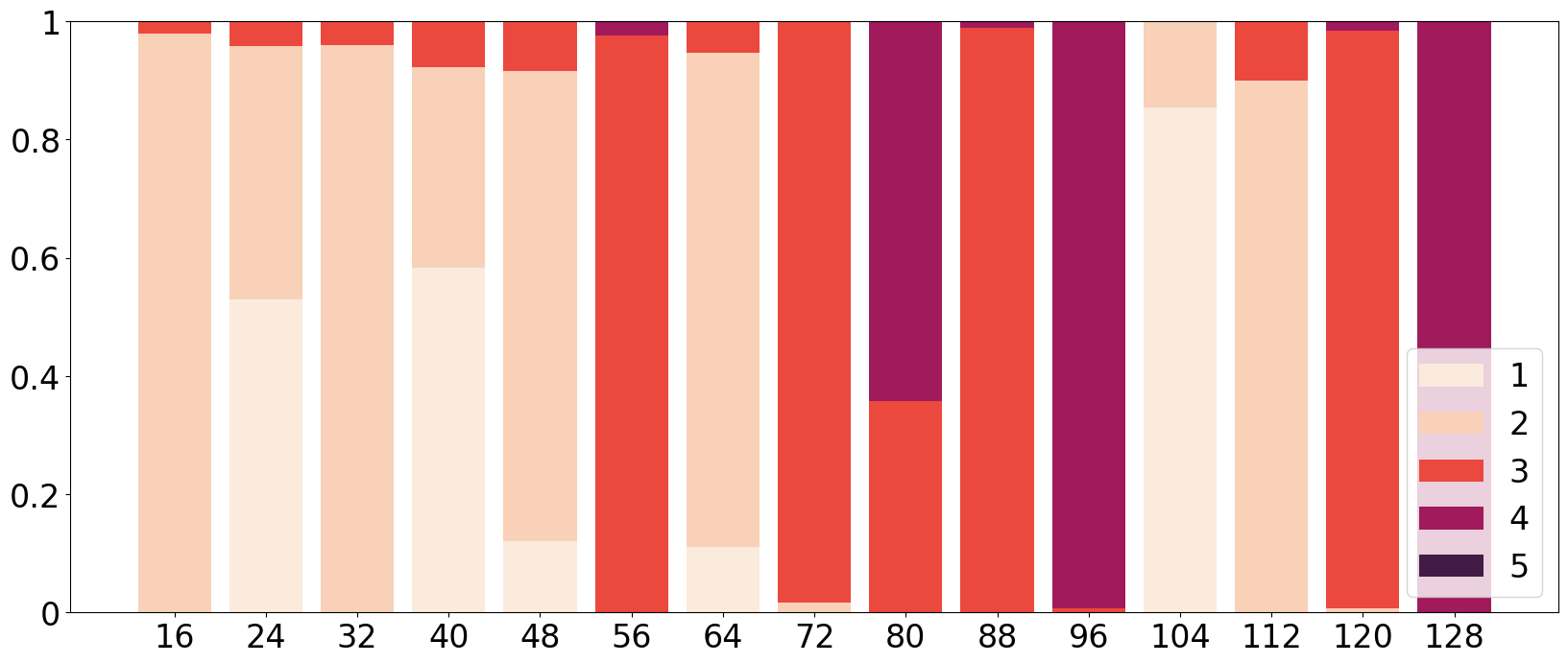}
\end{subfigure}
  \begin{subfigure}{.49\textwidth}
    \centering
    \includegraphics[width=1\linewidth]{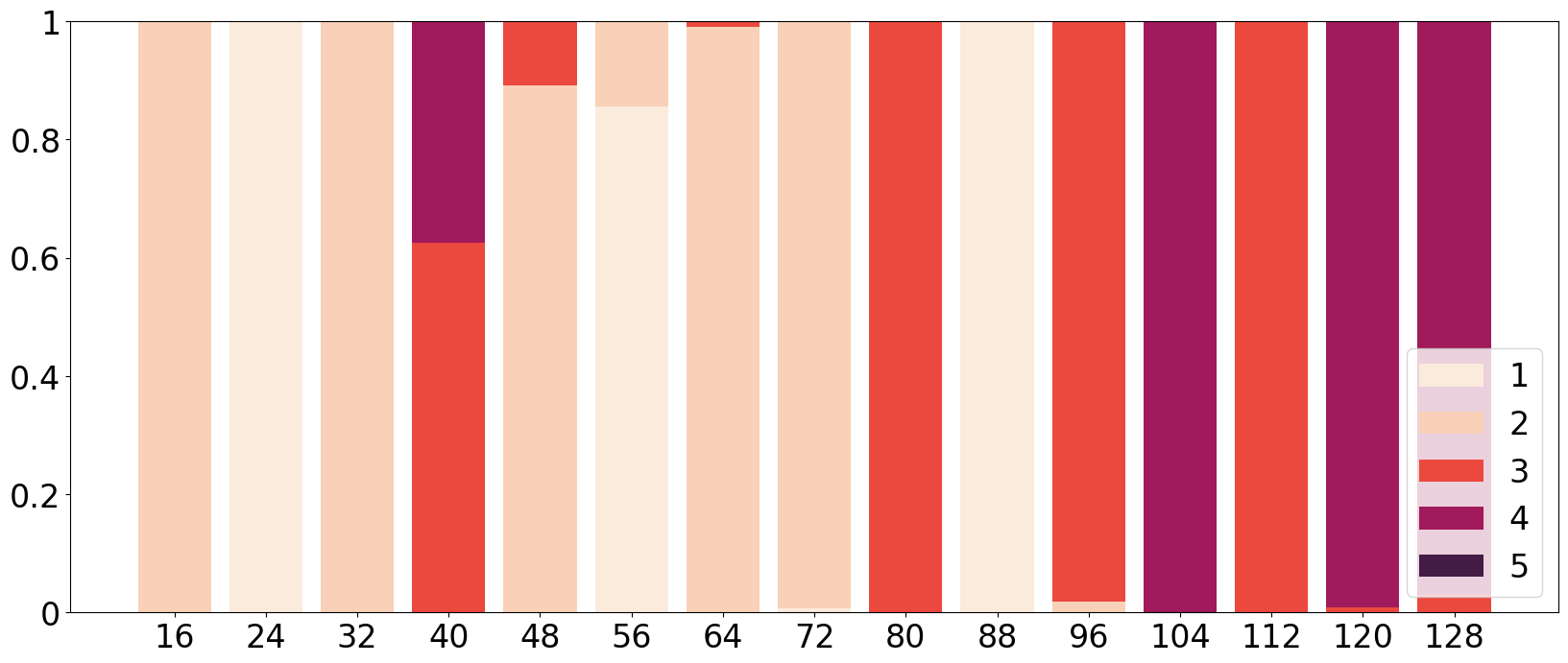}
\end{subfigure}
\begin{subfigure}{.49\textwidth}
    \centering
    \includegraphics[width=1\linewidth]{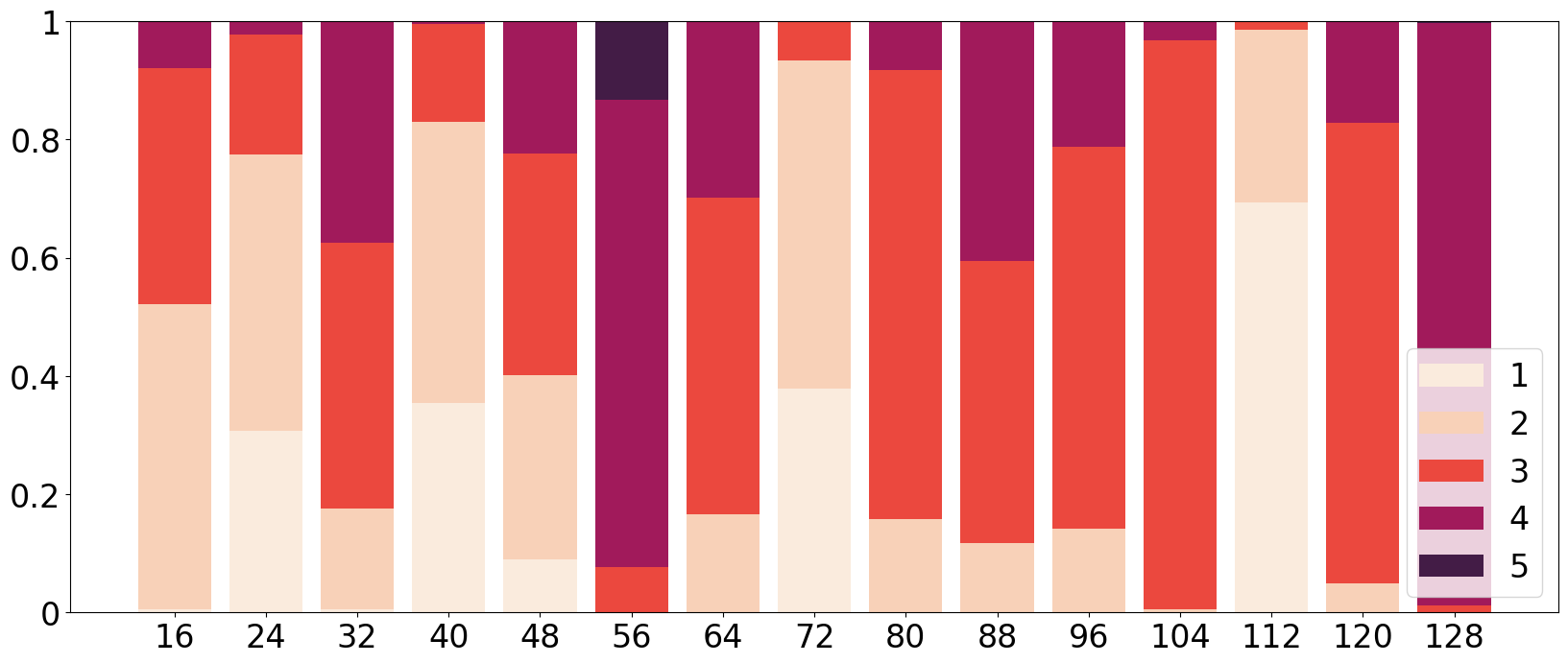}
\end{subfigure}
  \begin{subfigure}{.49\textwidth}
    \centering
    \includegraphics[width=1\linewidth]{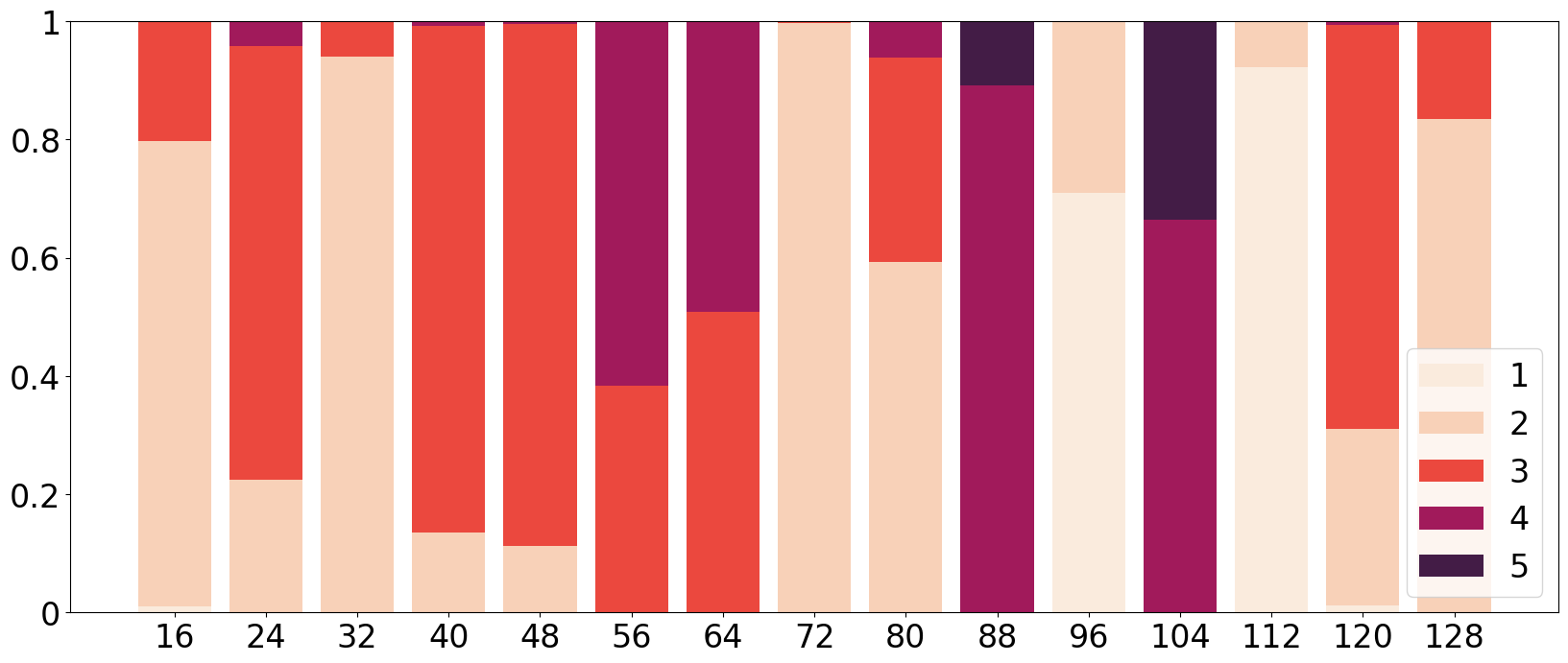}
\end{subfigure}
\begin{subfigure}{.49\textwidth}
    \centering
    \includegraphics[width=1\linewidth]{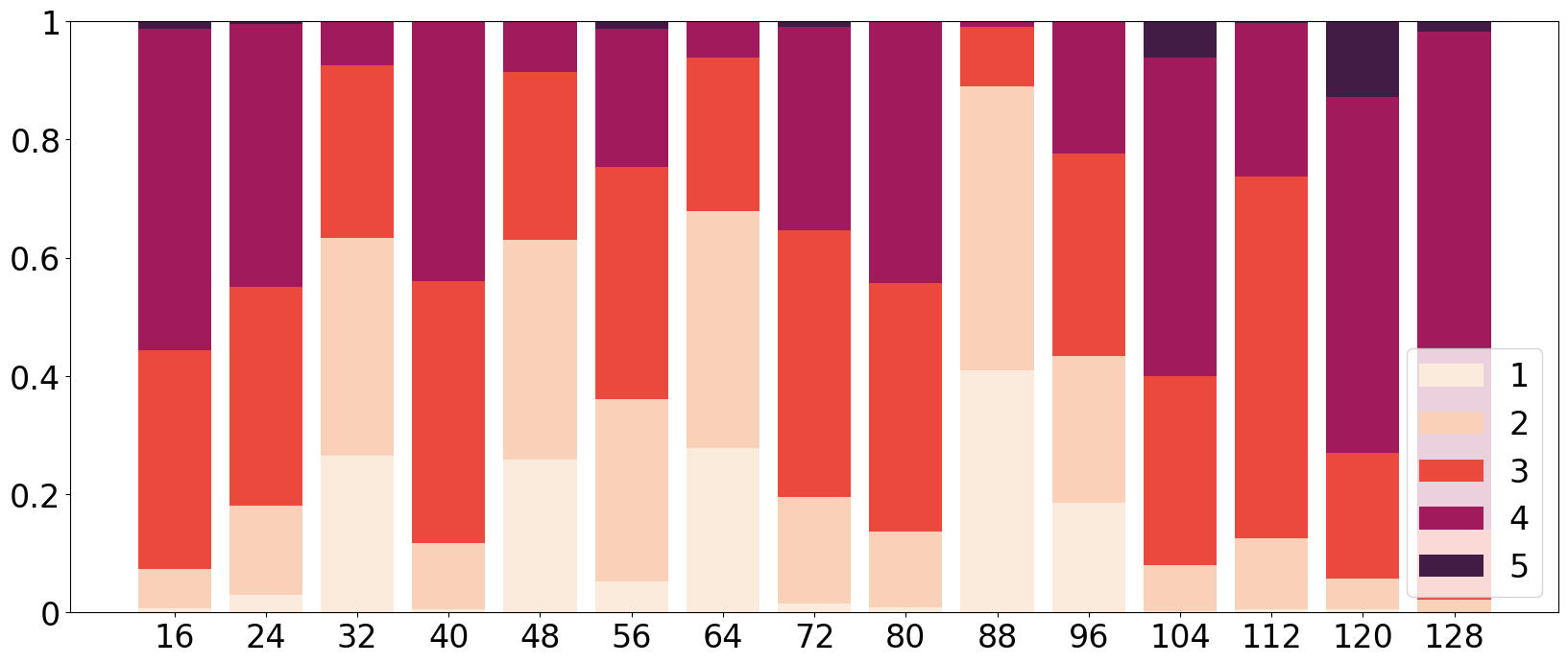}
\end{subfigure}
\begin{subfigure}{.49\textwidth}
    \centering
    \includegraphics[width=1\linewidth]{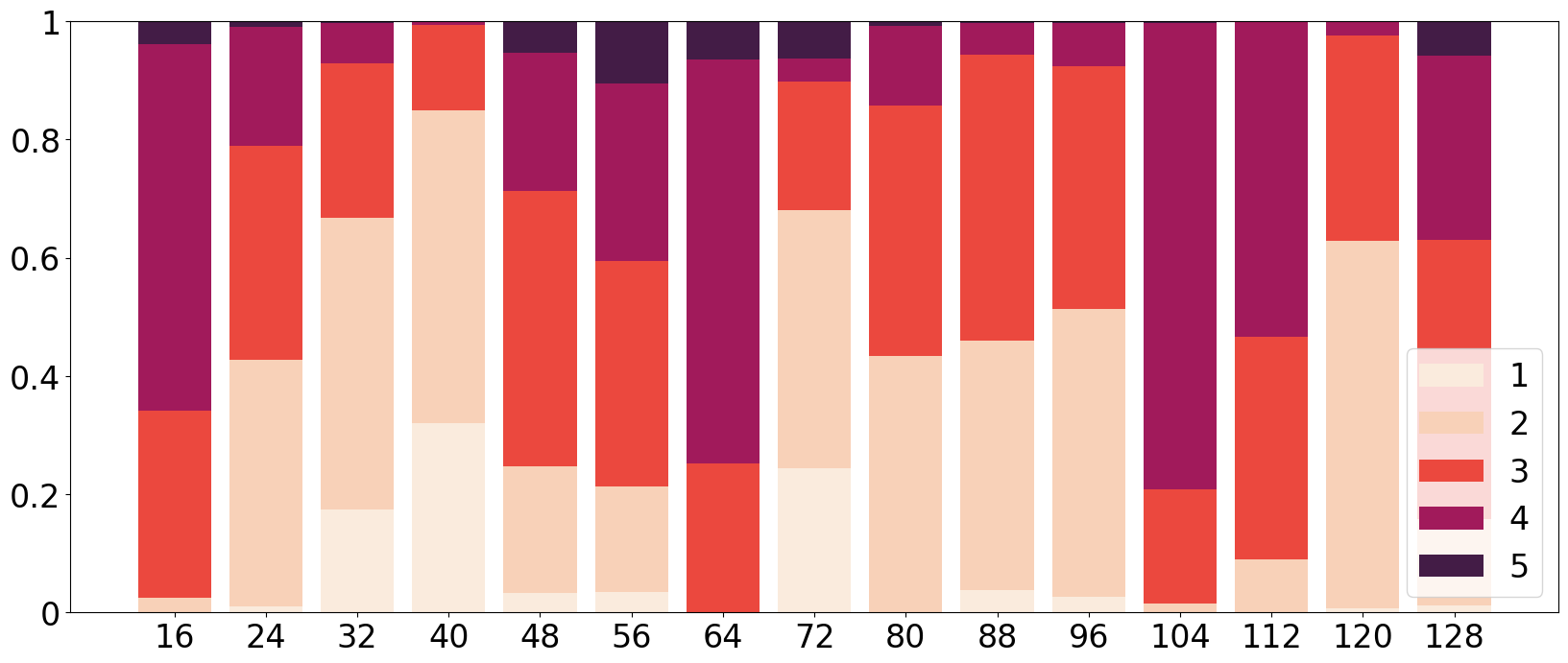}
\end{subfigure}
  \vspace{-5pt}
\caption{
    The ratio of an expert activated by different numbers of tasks in all the MLoRE modules in our model  
    when the task-sharing generic path is not equipped.
    Horizontal coordinates represent the ranks of different experts.
  }\label{fig:vis_relationb_other}
  \vspace{-5pt}
\end{figure*}

We present the relation visualization between tasks and the low-rank experts from 
every scale in \figref{fig:vis_relationa_other}.
These visualizations clearly show that experts with different ranks tend to learn 
different subsets of tasks for all the MLoRE modules.
We also show the ratio for different experts to be activated by different numbers of tasks when the task-sharing generic path is not added for all the MLoRE modules in \figref{fig:vis_relationb_other}.
Most of the experts in MLoRE modules are seldom activated by all the tasks which ties 
in with our motivation.
A few experts are shared by all the tasks more frequently, though, we find that the minimum 
gating value among the five tasks for these experts is relatively low, 
which is under half of the average value (0.11) for most of the time.
This can also prove that it is hard for one expert to build global relationships 
and effectively aid the final results without the task-sharing generic path.

{\small
\bibliographystyle{ieee_fullname}
\bibliography{mtl}
}

\end{document}